# Duck swarm algorithm: theory, numerical optimization, and applications

Mengjian Zhang[1] · Guihua Wen[1]

**Abstract**
A swarm intelligence-based optimization algorithm, named Duck Swarm Algorithm (DSA), is proposed in this study, which is inspired by the searching for food sources and foraging behaviors of the duck swarm. Two rules are modeled from the finding food and foraging of the duck, which corresponds to the exploration and exploitation phases of the proposed DSA, respectively. The performance of the DSA is verified by using multiple CEC benchmark functions, where its statistical (best, mean, standard deviation, and average running-time) results are compared with seven well-known algorithms like Particle swarm optimization (PSO), Firefly algorithm (FA), Chicken swarm optimization (CSO), Grey wolf optimizer (GWO), Sine cosine algorithm (SCA), and Marine-predators algorithm (MPA), and Archimedes optimization algorithm (AOA). Moreover, the Wilcoxon rank-sum test, Friedman test, and convergence curves of the comparison results are utilized to prove the superiority of the DSA against other algorithms. The results demonstrate that DSA is a high-performance optimization method in terms of convergence speed and exploration–exploitation balance for solving the numerical optimization problems. Also, DSA is applied for the optimal design of six engineering constrained optimization problems and the node optimization deployment task of the Wireless Sensor Network (WSN). Overall, the comparison results revealed that the DSA is a promising and very competitive algorithm for solving different optimization problems.

## 1 Introduction

Optimization algorithms play a significant role in solving the real-world optimization problems. Especially, these algorithms can be compartmentalized different categories using different descriptions. Common names are evolutionary algorithm (EA) [1], nature-inspired algorithm (NIA) [2], meta-heuristic algorithm (MA) [3], and swarm intelligence (SI) algorithm [4], however, some of the algorithms included are the same. Thus, a challenge of the algorithm is that searching for the optima in the search space with higher convergence speed. Three typical and noted heuristic algorithms (evolutionary algorithms), Genetic Algorithm (GA) [5], Simulated Annealing (SA) [6] algorithm, and Particle Swarm Optimization (PSO) algorithm [7], have made great contributions and provided a lot of reference for the algorithms that were proposed later.

**Genetic Algorithm** combines evolution and natural selection, which are applied to its population over generations, and it was proposed in the 1970s [5, 8]. The best chromosomes in the previous generation or generated by crossover and mutation constitutes the next population in the optimization process. The crossover is to inherit a part of the value of two chromosomes from each parent and produces one offspring, which can direct to the exploitation. The mutation is randomly changing some values in a chromosome and responsible for the exploration. Overall, highly random operations make GA avoid falling into local optimum, and slow convergence is its disadvantage at the same time.

**Simulated Annealing** [6] was proposed in 1983, one of the most well-known physics-based methods, which is inspired by the annealing in metallurgy. It starts to find the global optimal solution at a high "temperature" and becomes more sensitive as the temperature decreases, that is, the ratio of the difference solution decreases. Thus, the initial temperature and annealing speed are the key indicators that affect whether it can reach the optimum.

✉ Mengjian Zhang
mjz960106@163.com

[1] School of Computer Science and Engineering, South China University of Technology, Guangzhou 510006, China

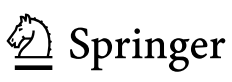

**Particle Swarm Optimization** (PSO) algorithm was proposed in 1995 [7], one of the most popular SI methods, which inspired by the bird flocking behavior. The movements of the particles are affected by the position and speed of the previous generation and the surrounding particles. PSO algorithm has a clearer direction than GA and SA, because it is easy to implement, and the parameters are rarely the outstanding advantages of the PSO. However, it tends to converge to the local optimum prematurely when optimizing multi-modal functions, because it uses the static finite predecessor and group of linear motion. Above the three methods, their variations have been proposed, such as Quantum PSO [9], Adaptive PSO [10], and Hybrid GA with SA [11], etc.

During the last two decades, many meta-heuristic algorithms were proposed and have been used for solving optimization problems after GA, SA, and PSO. Some of the most well-known optimization techniques are Differential evolution (DE) [12], Harmony search (HS) [13], Ant colony optimization (ACO) [14], Firefly algorithm (FA) [15], Cuckoo search (CS) [16], Gravitational search algorithm (GSA) [17], Grey wolf optimizer (GWO) [18]. To some extent, the algorithms mentioned above are inspired by some, such as the social behavior of animal groups (foraging, migration, courtship), the evolution of nature, human social behavior, etc. Thus, we can name all of them inspiration algorithm in this paper. These optimization algorithms have succeeded to solve optimization problems of the literature. However, according to the No Free Lunch (NFL) theorem [19] that no inspiration algorithm best for solving all optimization problems. Namely, this indicates that an inspiration algorithm may produce satisfying solutions on a set of problems but unsatisfying solutions on another set of problems. Thence, this motivates our essays to develop a novel swarm intelligence algorithm with inspiration from duck swarm.

This study proposes a novel swarm intelligence algorithm, named Duck Swarm Algorithm (DSA), for solving the numerical optimization functions from the CEC benchmark functions, the real-world engineering constrained optimization problems and the WSN's node optimization deployment task. The inspirations behind the proposed algorithm are the search and foraging behaviors of the duck swarm. The main **contributions** are as follows:

- Inspired by the social behaviors of the duck swarm which summarized observations in daily life, a novel group-based swarm intelligence algorithm is proposed, named Duck Swarm Algorithm (DSA).
- Two rules are modeled from the finding food and foraging of the duck, which corresponds to the exploration and exploitation phases of the proposed DSA, respectively. Where duck cooperation and competition are considered in details.
- The proposed DSA performs well on multiple CEC benchmark functions, and outperforms for solving the engineering constrained optimization problems and the WSN node optimization deployment task.

The rest is set up as follows: Sect. 2 reviews the literature of the nature-inspired metaheuristic algorithms. Section 3 introduces the proposed DSA in detail. Section 4 presents the comparison experiments of the algorithms. Experiments and simulations of the DSA's performance are described, and the results are illustrated in separate graphical diagrams in Sect. 5. Moreover, the conclusion and future work are discussed in Sect. 6.

## 2 Literature review

According to the inspiration principle of the meta-heuristic optimization algorithms, which can be simply categorized into four categories (See Fig. 1). Based on the inspiration source, the mainly four classes are: (i) evolution-based algorithms, (ii) swarm intelligence algorithms, (iii) physics-based algorithms, and (iv) human behavior-based algorithms. Of course, all meta-heuristic optimization methods benefit from these advantages despite the differences.

The first main division of meta-heuristics is evolution-based methods. Such evolutionary algorithms normally mimic evolutionary rules in nature, some of the most well-known techniques are Genetic algorithm (GA) [8], Genetic programming (GP) [20], Differential evolution (DE) [12], Evolutionary programming (EP) [21], Biogeography-based optimizer (BBO) [22], Gradient evolution algorithm (GEA) [23], and Tree-seed algorithm (TSA) [24].

The second main division of meta-heuristics is swarm-based approaches. These SI algorithms currently mimic swarm behaviors in animals. Some of the most popular algorithms are Particle swarm optimization (PSO) [7], Ant colony optimization (ACO) [14], Firefly algorithm (FA) [15], Cuckoo search (CS) [25], Grey wolf optimizer (GWO) [18], Salp swarm algorithm (SSA) [26], and Marine-predators algorithm (MPA) [27]. Besides, some well-known SI algorithms are not listed in Fig. 1 that are Whale optimization algorithm (WOA) [28, 75] inspired by the foraging and hunting of the whales in the ocean, Moth-flame optimization (MFO) [29] inspired by the navigation approach of moths, and Butterfly optimization algorithm

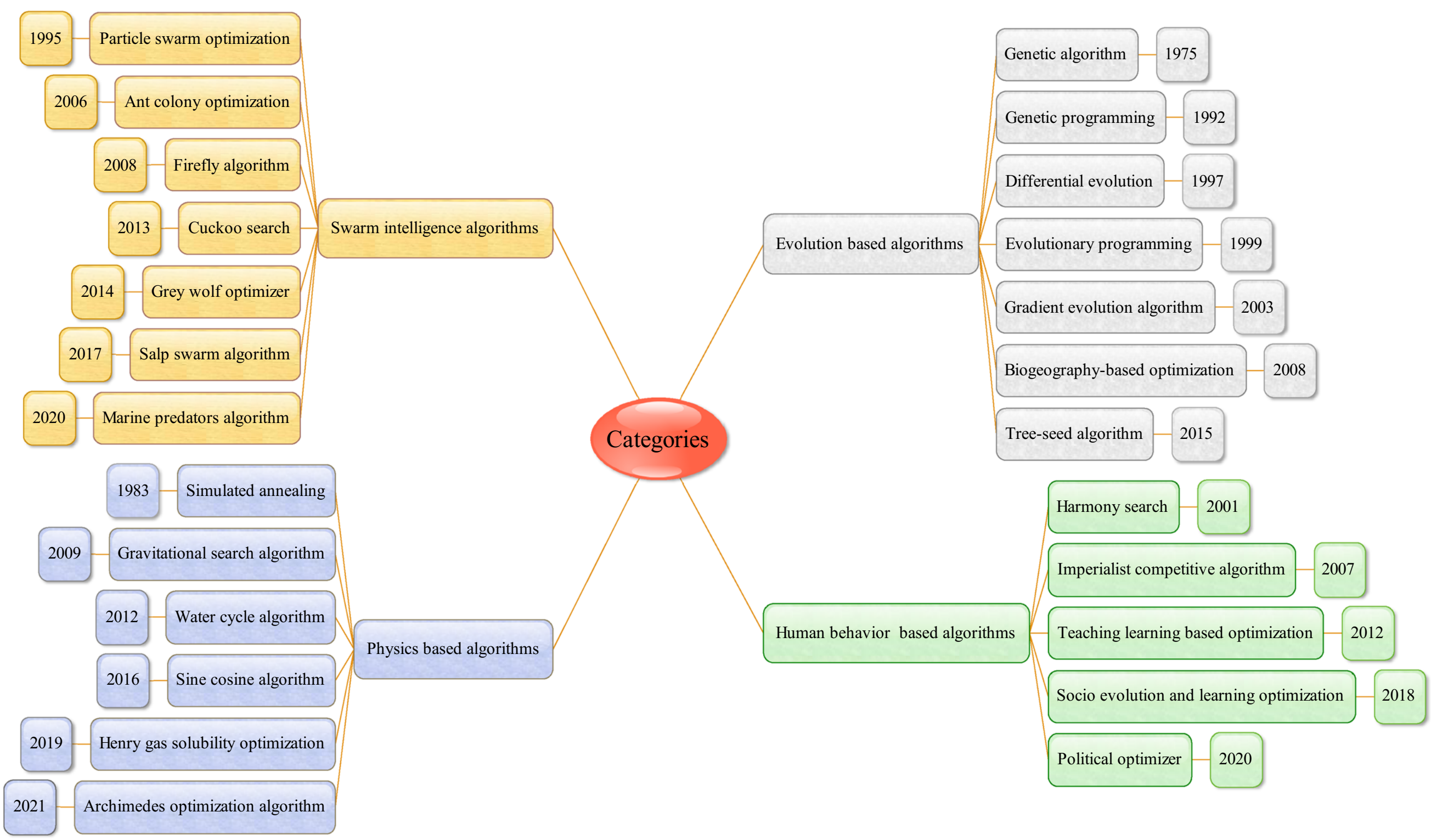

**Fig. 1** Classification of meta-heuristic optimization algorithms

(BOA) [30] inspired by the foraging and mating behaviors of butterflies, etc.

The third main division of meta-heuristics is physics-based methods. These optimization algorithms usually mimic physical principle. Some of the well-known methods are Simulated annealing (SA) [6], Gravitational search algorithm (GSA) [17], Water cycle algorithm (WCA) [31], Sine cosine algorithm (SCA) [32], Henry gas solubility optimization (HGSO) algorithm [33], and Archimedes optimization algorithm (AOA) [34]. It is worth mentioning that AOA was proposed in 2021 by Fatma et al., which inspired from the phenomenon explained by Archimedes' principle. Also, Equilibrium optimizer (EO) [35] and Gradient-based optimizer (GBO) [36] are proposed for solving the numerical optimization problems inspired by the physical rules.

The fourth main division of meta-heuristics is human social behavior-based tools. Such optimization algorithms typically mimic social behavior rules in humans. Some of the popular algorithms like Harmony search (HS) [13], Imperialist competitive algorithm (ICA) [37], Teaching learning-based optimization (TLBO) [38], Socio evolution and learning optimization (SELO) [39], and Political optimizer (PO) [40]. We divide HS algorithm into social behavior is based on the harmony that only humans can sing, and its principle include the description of the propagation of musical sound. For a more detailed review, different categories can refer to the literature [18, 41–43].

Overall, various SI methods have been proposed recently. Most of these approaches is inspired by foraging, mating, hunting and searching behaviors of animals in nature. In the scope of our knowledge, there is no SI method in the literature inspired by the social behaviors of duck swarm. This is the main motivation for proposing a new SI method by modeling the social behavior of the duck swarm. Additionally, its abilities for solving the numerical, real-world engineering optimization problems and the WSN node optimization deployment task are investigated in the following.

## 3 Duck Swarm Algorithm

In this section, a novel SI optimization algorithm, named Duck Swarm Algorithm (DSA), is proposed that imitates the searching for food and foraging behaviors of the duck swarm. To understand this new algorithm some biological facts and how to model them of the DSA are discussed in details below.

### 3.1 Inspiration

In nature, formation characteristics are common for group animals, especially in the process of animal migration and foraging (hunting). Among group mammals, there are also obvious hierarchical characteristics, such as: lions, wolves, monkeys, etc. The GWO algorithm proposed for the hierarchical system of the grey wolves, the algorithm divides the hunting characteristics of grey wolves into four levels. Moreover, there are many intelligent algorithms proposed for group animals belonging to birds, including classic PSO algorithm, CS algorithm, Crow search algorithm (CSA) [44], Chicken swarm optimization (CSO) algorithm [45], Sparrow search algorithm (SSA) [46], Honey Badger Algorithm (HBA) [58], etc.

Ducks are aquatic and terrestrial amphibians, and it can be simply divided into three common duck species [47]: water ducks, diving ducks, and roosting ducks. The common duck (poultry) in our life belongs to the water duck and the order Anseriformes. It is generally considered to be a bird. The nature-inspired heuristic algorithms are derived from the observation of phenomena, such animals, plants, or other characteristics in nature. Then, their behaviors are abstracted into mathematical models, and designed as optimization methods for solving numerical optimization problems, and engineering constraint optimization problems, and other real-world optimization problems.

It can be seen from observation that duck swarm queuing, searching for food sources and foraging behaviors have certain laws in life. Some pictures of duck swarm behaviors are provided in Fig. 2.

## 3.2 Mathematical model of DSA

This section detailed presents the mathematical model of the proposed approach. Three main processes of the DSA are discussed as follows: (i) Positions of duck swarm after queuing (Population initialization), (ii) Searching for food sources (Exploration phase), (iii) Foraging in groups (Exploitation phase). Noted that there are two rules that need to be followed in the process of searching food of ducks. Rule one: when looking for food, ducks with strong search ability are located closer to the center of food source, which attract other individuals to move closer to them; the updated location is also affected by nearby individuals. Rule two: when foraging, the individuals all approach the food; the next position is affected by neighboring individuals and food position or leader duck.

### 3.2.1 Population initialization

Supposing the expression of randomly generated initial position in the $D$-dimensional search space is as follow:

$$X_i = L_b + (U_b - L_b) \cdot o \tag{1}$$

where $\boldsymbol{X}_i$ denotes the spatial position of the $i$-th duck ($i$ = 1, 2, 3, …, $N$) in the duck group, $N$ is the number of population size. $\boldsymbol{L}_b$ and $\boldsymbol{U}_b$ represent the upper and lower bounds of the search space, respectively. $\boldsymbol{o}$ is a random number matrix between (0, 1).

### 3.2.2 Exploration phase

After the queuing behavior of duck swarm, that is, the ducks arrived at a place with more food. Each-individual gradually disperses and starts searching for food, this process is defined as follows:

$$X_i^{t+1} = \begin{cases} X_i^t + \mu \cdot X_i^t \cdot sign(r - 0.5), if\, P > rand \\ X_i^t + CF_1 \cdot (X_{leader}^t - X_i^t) + CF_2 \cdot (X_j^t - X_i^t), if\, P \leqslant rand \end{cases} \tag{2}$$

where sign($r$-0.5) influences the process of searching for food, and it can be set either -1 or 1.$\mu$ denotes the control parameter of global search. $P$ is searching conversion probability of exploration phase. $CF_1$ and $CF_2$ denote cooperation and competition coefficient between ducks in the search stage, respectively. $X_{leader}^t$ represents the best duck position of the current historical value in the $t$-th iteration. $X_j^t$ denotes the agents around $X_i^t$ in searching for food by duck group in the $t$-th iteration. Moreover, parameter $\mu$ can be calculated as follows:

$$\mu = K \cdot (1 - t/t_{\max}) \tag{3}$$

where $K$ is calculated by:

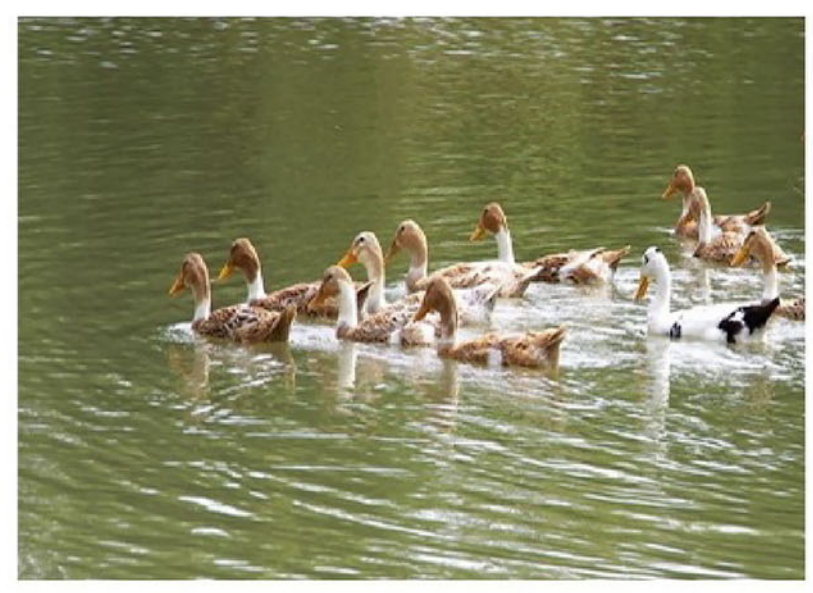

(a) Queuing behavior of duck swarm

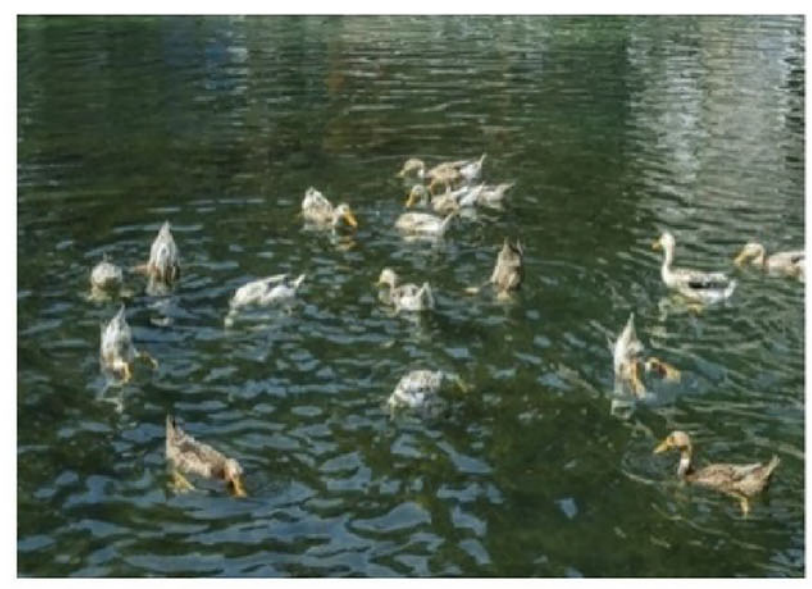

(b) Searching for food sources

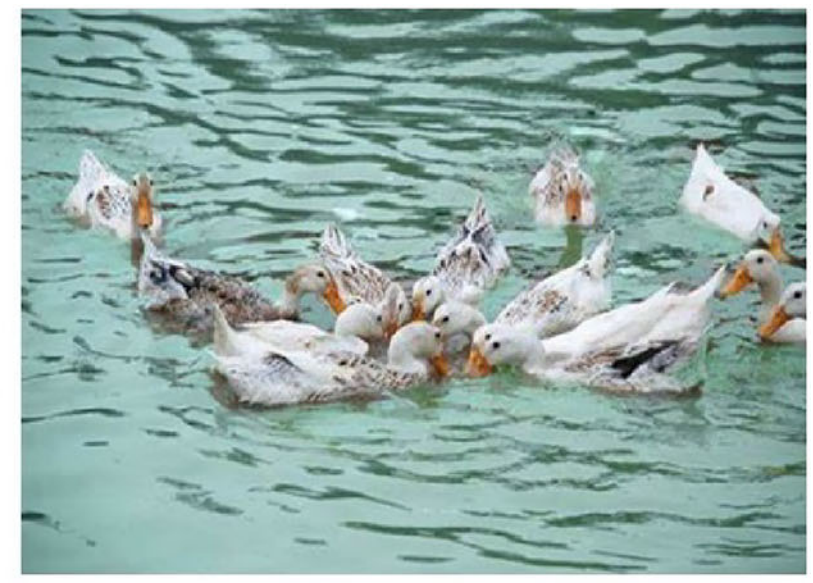

(c) Foraging in groups

**Fig. 2** Behaviors of the duck swarm

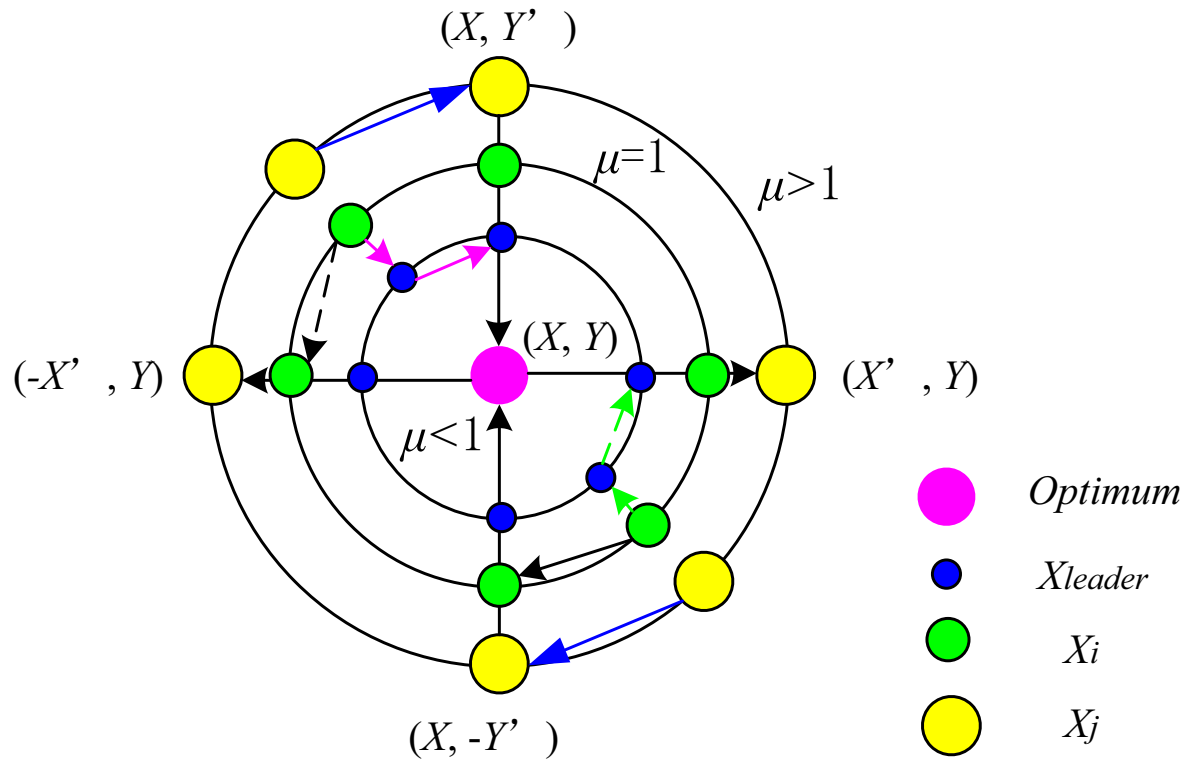

(a) Sketch map of the exploration phase

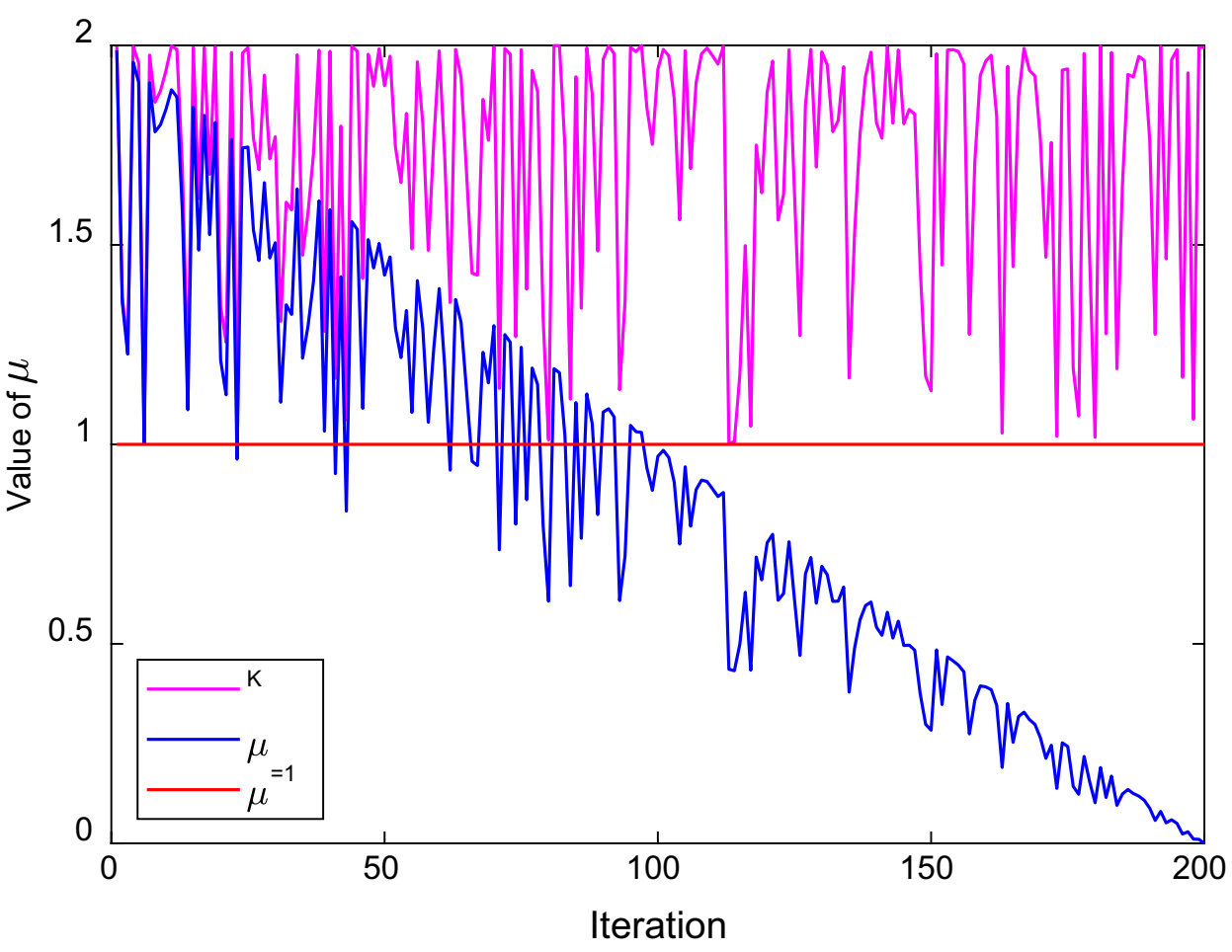

(b) The curves of parameters $K$ and $\mu$

**Fig. 3** Sketch map of the exploration phase and parameters $K$ and $\mu$ curves of the proposed DSA

$$K = \sin(2 \cdot rand) + 1 \tag{4}$$

In the exploration phase, Fig. 3a depicts the process of agents update its position pertaining to $X_i$, $X_j$, and $X_{leader}$ in a 2-D search space. The value curves of parameters $K$ and $\mu$ with 200 iterations are shown in Fig. 3b.

As shown in Fig. 3, the search range of duck swarm is wider when $\mu > 1$ in the exploration phase. This non-linear strategy is used to enhance the global search ability of the proposed DSA. Besides, this phase is also can be integrated, and two parameters $C_1$ and $C_2$ will be considered in the updating formula, which are used to balance the positions of the individual in the exploration phase. Notably, the integrated formula is as follows:

$$\begin{aligned} X_i^{t+1} = X_i^t + C_1 \cdot \mu \cdot (X_i^t - C_2 \cdot X_{leader}^t) \cdot sign(r - 0.5) \\ + CF_1 \cdot (X_{leader}^t - X_i^t) + CF_2 \cdot (X_j^t - X_i^t) \end{aligned} \tag{5}$$

where $\mu$ denotes the control parameter of global search. $C_1$ and $C_2$ are the position balance adjustment parameters, $C_1$ is a random number in (0, 1). Notably, $C_2$ can not only be set as a random, but also as a constant, like 0, 1 or others. $CF_1$ and $CF_2$ denote cooperation and competition coefficient between ducks in the search stage, respectively. $X_{leader}^t$ indicates the best duck position of the current historical value in the $t$-th iteration. $X_j^t$ represents the individuals around $X_i^t$ in searching for food by duck group in the $t$-th iteration.

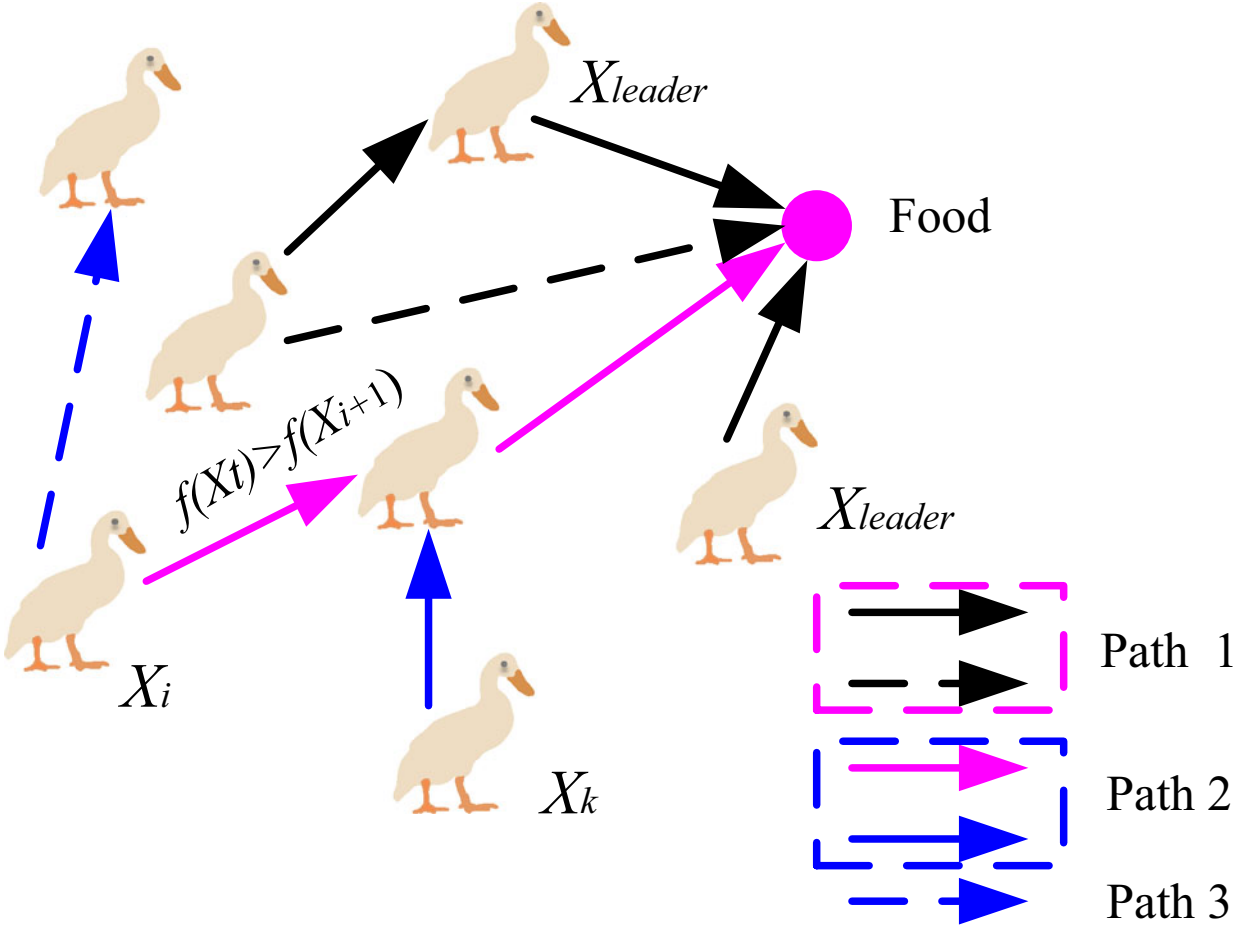

**Fig. 4** Sketch map of exploitation phase

### 3.2.3 Exploitation phase

After the duck swarm searching for enough food, which can satisfy the foraging of the ducks. This process is closely related to fitness of each duck's position and defined as follows:

$$X_i^{t+1} = \begin{cases} X_i^t + \mu \cdot (X_{leader}^t - X_i^t), if f(X_i^t) > f(X_i^{t+1}) \\ X_i^t + KF_1 \cdot (X_{leader}^t - X_i^t) + KF_2 \cdot (X_k^t - X_j^t), otherwise \end{cases} \tag{6}$$

where $\mu$ denotes the control parameter of global search in exploitation phase; parameters $KF_1$ and $KF_2$ denote the cooperation and competition coefficient between ducks in the exploitation phase, respectively. $X_{leader}^t$ represents the best duck position of the current historical value in the $t$-th iteration. $X_k^t$ and $X_j^t$ denote the agents around $X_i^t$ in foraging of duck group in the $t$-th iteration, where $k \neq j$.

Noted that the values of parameters $CF_1$, $CF_2$, $KF_1$ and $KF_2$ are all in (0, 2), and the calculation formula can be summarized as follows:

$$CF_i \, or \, KF_i \leftarrow \frac{1}{FP} \cdot rand(0, 1)(i = 1, 2) \tag{7}$$

where $FP$ is a constant, it is set to 0.618; the $rand$ is a random number in (0, 1).

In the exploitation phase, Fig. 4 depicts the process of ducks update its position pertaining to $X_i$, $X_j$, $X_k$, and $X_{leader}$

in a 2-D search space. Path 1 denotes the choice of ducks with cooperation. Path 2 represents the competition between $X_i$, $X_k$ and $X_j$ in the $t$-th iteration. Path 3 indicates the choice of the duck that have failed to compete the food search process.

## 3.3 Pseudo-code and complexity analysis of the DSA

### 3.3.1 Pseudo-code of the DSA

A detailed algorithm optimization process needs to be demonstrated, in which, the designed DSA theory is introduced in the Sect. 3.2. The pseudo-code of the DSA is listed in

**Algorithm 1** Pseudo-code of the DSA

1. Input: Initial parameter value setting; population number $N$; initial positions of duck swarm; objective function
2. Calculate the fitness value of initial positions; and select the best value $f_{\min}$ and leader agent position $X_{leader}$
3. While $t < T_{\max}$
4. Update the value of parameter $\mu$ using Eq. (3); and update the parameters $P$, $KF_1$, $KF_2$, $KF_1$ and $KF_2$
5. For $i$=1: size ($N$) **%Exploration phase**
6. Update the positions of duck swarm using Eq. (2) or Eq. (5)
7. Determine whether the individual is out of the search range
8. Calculate the new position and fitness value $f_{\text{new}}$
9. Update the leader position $X_{leader}$ and fitness value
10. End For
11. For $i$=1: size ($N$) **%Exploitation phase**
12. Update the positions of duck swarm using Eq. (6)
13. Determine whether the individual is out of the search range
14. Calculate the new fitness value $f_{\text{new}}$
15. if $f_{\text{new}}$ < fitness
16. Update the individual's position and fitness value
17. end if
18. Update the leader position $X_{leader}$ and fitness value
19. End For
20. End While
21. Output: the best position and fitness value

Besides, the flowchart of the designed DSA is presented in Fig. 5. The phases of the Exploration (Stage 2) and Exploitation (Stage 3) are shown in details, which corresponds to the pseudo-code of the proposed DSA.

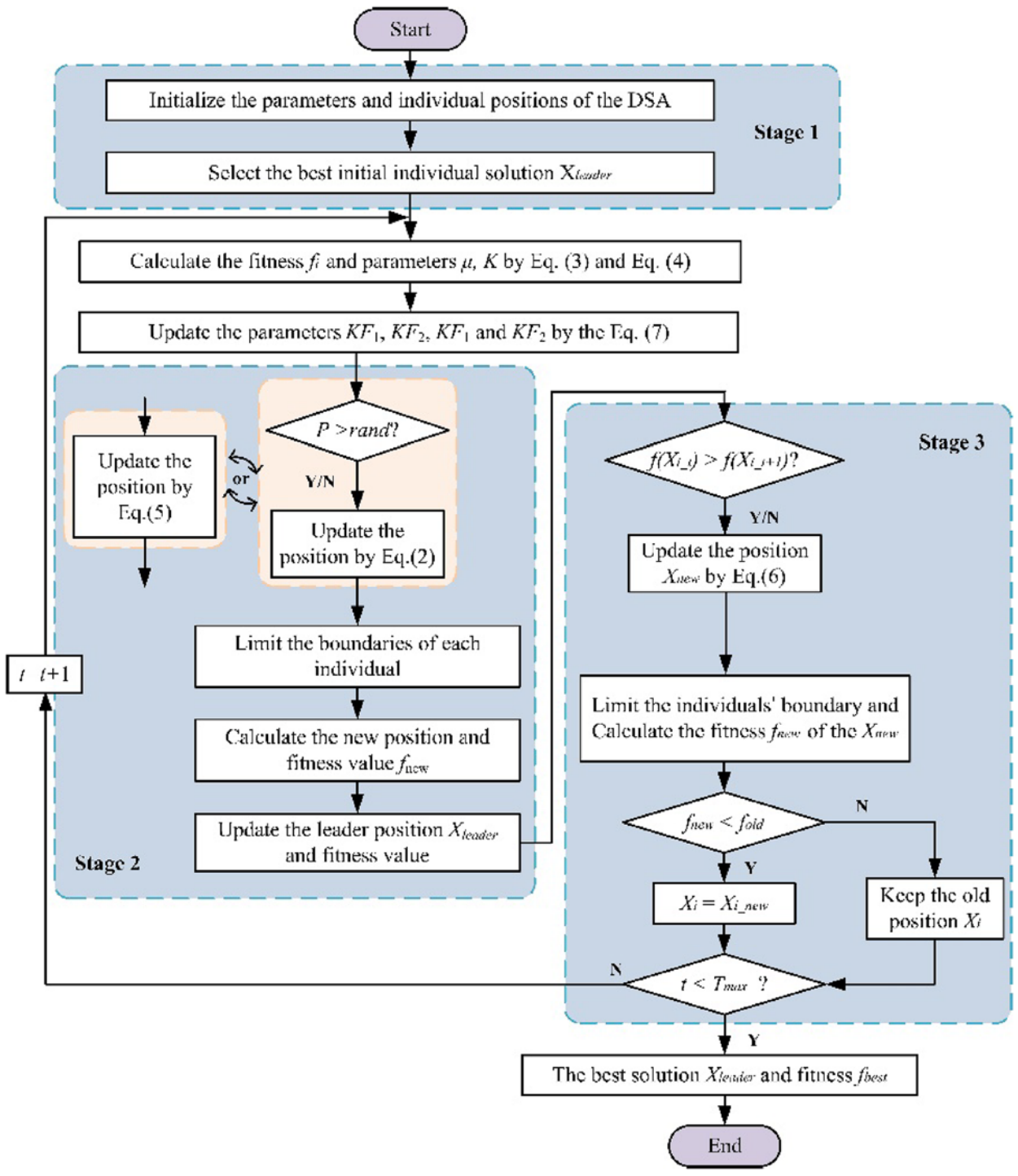

Fig. 5 Flowchart of the designed DSA

### 3.3.2 Complexity analysis

In this subsection, time and space complexity of the DSA are presented.

**3.3.2.1 Time complexity** Assuming that the population size and the search space dimension of the problem are $n$ and $d$, and the maximum iteration is $T$. The complexity of the DSA includes: the population initialization complexity is $O(nd)$, the fitness value of calculation complexity is $O(nd)$, the exploration and exploitation phases update complexity are $O(T)(n + nlogn + n + nlogn)$, and the parameters update complexity of the method is $O(T)$. To the above parts, the total time complexity of the proposed DSA is expressed as:

$$O(DSA) = O(nd) + O(T)O(1 + 2nd + 2nlogn) \tag{8}$$

**3.3.2.2 Space complexity** The storage space consumed by an algorithm can be defined the space complexity. It is closely related to the population size ($n$) of the algorithm and the dimension ($d$) of the problem. The total space complexity of the proposed DSA is $O(n{\cdot}d)$. Thus, the space efficiency of the proposed method is effective and stable.

**Table 1** Parameters of comparison algorithms

| Algorithms | Parameters | Value |
|---|---|---|
| PSO | Max and min velocity of particles | −1, 1 |
| | Cognitive and social constants | 2, 2 |
| | Inertial weight | Linearly decreases from 0.9 to 0.2 |
| FA | Alpha, beta, and gamma | 0.2, 1, 1 |
| CSO | Parameter *G* and *FL* | 10, [0.5, 0.9] |
| GWO | Parameter *a* | Linearly decreases from 2 to 0 |
| SCA | Parameter *a* | 2 |
| MPA | Fish Aggregating Devices, FADs | 0.2 |
| AOA | $C_1$, $C_2$, $C_3$, $C_4$ | 2, 6, 2, 4 |
| DSA | $CF_1$, $CF_2$ | Random values in (0, 2) |
| | Parameters *P* and *FP* | 0.5 and 0.618 |
| | $KF_1$, $KF_2$ | Random values in (0, 2) |

## 4 Comparative experiment and statistical test design

To verify the proposed algorithm's efficiency, DSA has been compared to seven optimization algorithms. The comparison techniques are Particle swarm optimization (PSO, 1998) [48], Firefly algorithm (FA, 2008) [15], Chicken Swarm Optimization (CSO, 2014) [45], Grey wolf optimizer (GWO, 2014) [18], Sine cosine algorithm (SCA, 2016) [32], Marine-predators algorithm (MPA, 2020) [27], and Archimedes optimization algorithm (AOA, 2021) [34]. The initial parameter values of the seven competitive methods are listed in Table 1. Three categories of seven comparison algorithms are used to assess the DSA efficiency.

### 4.1 Comparative experiment

All the experimental series are carried out a Windows 10 system using Intel (R) Core (TM) i5-10210U CPU @2.11G with 8G RAM, and MATLAB 2018a in this study. For the statistical results like Mean, and Standard deviation (Std), the comparison algorithms performed 30 independent runs for each test function. The agent size in the population *N* is set to 30, and the max iteration of the comparison algorithms is set to 200. Additionally, the dimension of unimodal and multimodal functions is set to 30, which are selected from the CEC benchmark functions.

Eighteen functions [1, 17, 42, 49] from the IEEE CEC benchmark functions are used to assess the performance of the DSA in this study. Three groups of the test functions are unimodal, multimodal, and fixed-dimension numerical optimization problems. These functions are shown in the Table 2, including Search range, Dim dimension (*Dim*) of the function, and $f_{\min}$ is the optimum of the function in theory. Additionally, the 2-D versions of each benchmark function are illustrated in Fig. 6, where F1 ∼ F7, F8 ∼ F14, and F15 ∼ F18 are the 2-D versions of the unimodal, multimodal, and fixed-dimension problems, respectively.

### 4.2 Statistical test

The DSA is assessed via eighteen benchmark functions and compared with seven optimization algorithms. The performance of the DSA is evaluated by different test functions. The exploitation ability of competitive methods can be assessed by the unimodal functions because of only one global optimum of them (F1–F7). The exploration ability of competitive algorithms can be evaluated by the multimodal functions (F8–F14) because they have many local optima. The local optima avoidance ability between exploration and exploitation of the competitive algorithms can be assessed by fixed-dimension functions (F15–F18) as they have lots of local optima. The statistical results include the Best, Mean, Standard deviation (Std) of the optimal results with 30 times, and the average running time (Time/s) is also considered. They can be calculated as follows:

(1) The best value (Best)

$$F_{best} = \min(F_1, F_2, \ldots F_m) \tag{9}$$

where *m* indicates the number of optimization tests, and $F_{best}$ represents the optima in 30 independent runs.

(2) The mean value (Mean)

$$F_{mean} = \frac{1}{m}\sum_{i=1}^{m} F_i \tag{10}$$

where $F_i$ indicates the optimum in each independent run, and $F_{mean}$ denotes the mean value of the 30 independent runs.

(3) The Standard deviation (Std)

**Table 2** Unimodal benchmark functions

| Type | No | Function name | Search range | *Dim* | $f_{min}$ |
|---|---|---|---|---|---|
| Unimodal | F1 | Sphere | [− 100,100] | 30 | 0 |
| | F2 | Schwefel 2.22 | [− 10,10] | 30 | 0 |
| | F3 | Schwefel 1.2 | [− 100,100] | 30 | 0 |
| | F4 | Schwefel 2.21 | [− 100,100] | 30 | 0 |
| | F5 | Rosenbrock | [− 30,30] | 30 | 0 |
| | F6 | Cigar | [− 100,100] | 30 | 0 |
| | F7 | Quartic | [− 1.28,1.28] | 30 | 0 |
| Multimodal | F8 | Schwefel 2.26 | [− 500,500] | 30 | $-418.9829 \times Dim$ |
| | F9 | Rastrigin | [− 5,5] | 30 | 0 |
| | F10 | Ackley | [− 32,32] | 30 | 0 |
| | F11 | Griewank | [− 600,600] | 30 | 0 |
| | F12 | Penalized 1 | [− 50,50] | 30 | 0 |
| | F13 | Penalized 2 | [− 50,50] | 30 | 0 |
| | F14 | Weierstrass | [− 1,1] | 30 | 0 |
| Fixed-dimension | F15 | Shekel's Foxholes | [− 65,65] | 2 | 1 |
| | F16 | Kowalik's | [− 5,5] | 4 | 0.00030 |
| | F17 | Six-hump camel back | [− 5,5] | 2 | − 1.0316 |
| | F18 | Branin | [− 5,5] | 2 | 0.398 |

$$F_{std} = \sqrt{\frac{1}{m}\sum_{i=1}^{m}(F_i - F_{mean})^2} \tag{11}$$

(4) The average running time (Time/s)

$$T_{mean} = \frac{1}{m}\sum_{i=1}^{m} T_i \tag{12}$$

where $T_i$ indicates the running time for each single optimization.

The statistical results (include Best, Mean, Standard deviation (Std), and average running time (Time/s)) are given in Table 3, 4, 5. The best results are highlighted in bold. For the statistical results of comparison algorithms, statistical tests are required to assess the performance of the DSA sufficiently according to Ref. [50]. The statistical tests (Wilcoxon's rank-sum [51], and Friedman rank test [52]) are needed to suggest a remarkable improvement of a new swarm intelligence algorithm in comparison to the other well-known SI algorithms to solve a particular optimization problem. Wilcoxon's rank-sum (WRS) is a classical non-parametric statistical test that has been performed and reached the 5% significance level. Generally, p-value $< 0.05$ is considered strong evidence against the null hypothesis. Besides, the Friedman rank test is used to evaluate the superiority of the proposed DSA for solving optimization problems.

### 4.3 Population diversity test

According to Ref. [53, 54], to distinguish the diversity of agents in the process of exploration and exploitation, it is necessary to visually analyze the diversity of the population for a new SI algorithm. To analyze the population diversity of the proposed DSA, the diversity [55] is defined as follows:

$$Div(t) = \sum_{j=1}^{D}\frac{1}{N}Div_j = \frac{1}{N}\sum_{j=1}^{D}Div_j \tag{13}$$

where $Div(t)$ indicates population diversity in iteration $t$, $t$ is the current iteration during the optimization process, $N$ represents the population size, and $D$ is the dimension of the problem.$Div_j$ is calculated [54] as follows:

$$Div_j = \frac{1}{N}\sum_{i=1}^{N}\sqrt{\sum_{j=1}^{D}(X_{ij} - \bar{X}_j)^2}\ \bar{X}_j = \frac{1}{N}\sum_{i=1}^{N}X_{ij} \tag{14}$$

where $\bar{X}_j$ represents the mean of current solutions on dimension $j$, $Div_j$ indicates mean population diversity on dimension $j$, and $X_{ij}$ represents current solutions. Thus, the exploration and exploitation percentage measurement of the search process can be defined as follows:

$$Exploration(\%) = \frac{Div_t}{Div_{\max}} \times 100\% \tag{15}$$

$$Exploitation(\%) = \frac{|Div_t - Div_{\max}|}{Div_{\max}} \times 100\% \tag{16}$$

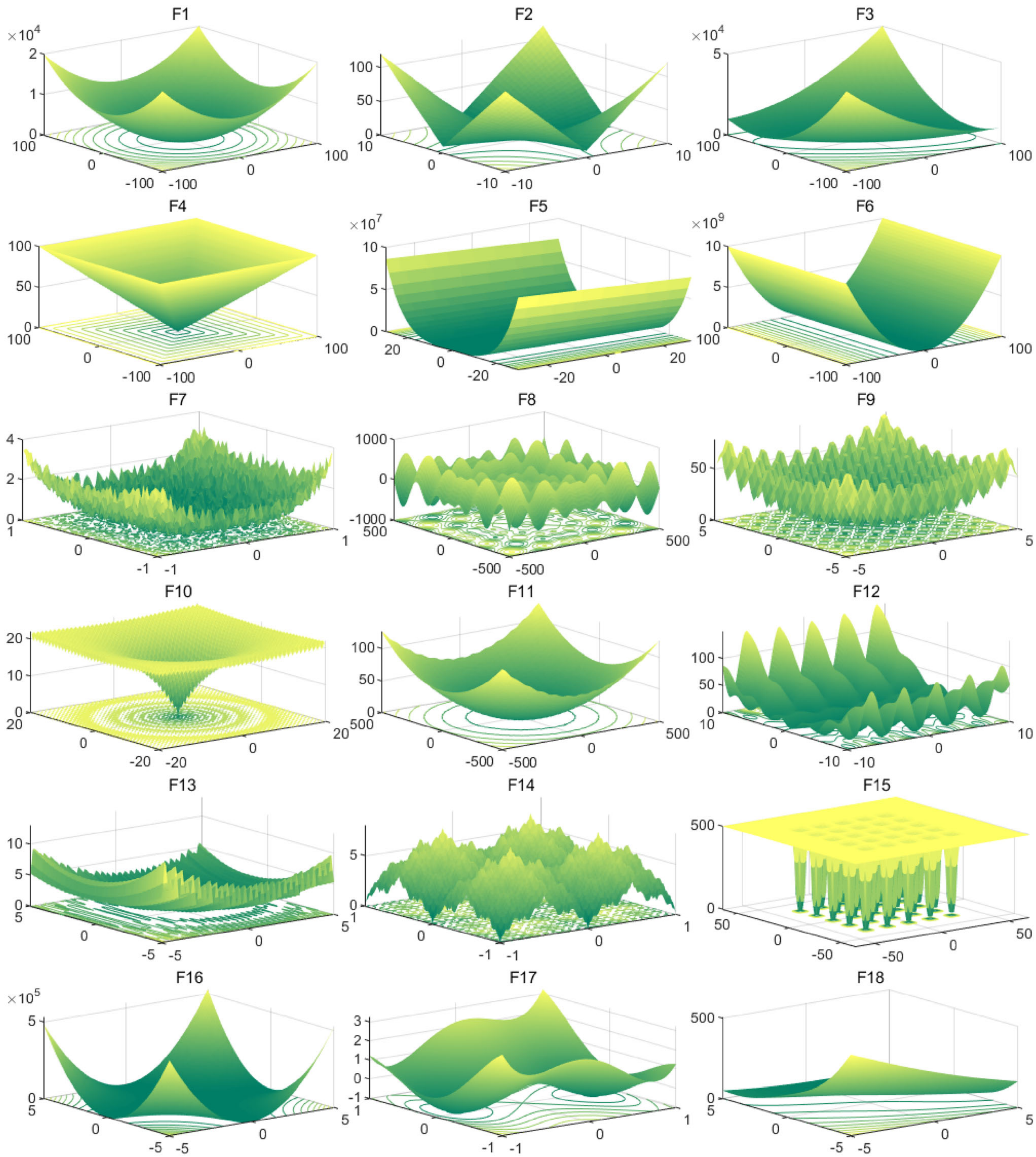

**Fig. 6** The 2-D versions of twenty-four benchmark functions

where $Div_t$ indicates population diversity of $t$-th iteration, and $Div_{\max}$ denotes the max diversity of the whole group's population diversity.

## 5 Results analysis and discussions

In this section, the experimental results of comparison algorithms are presented in Table 5, 6, 7. Figure 6 shows the convergence curves of the competitive algorithms for different types of functions. Figures 6 and 7 present the

**Table 3** Comparison statistical results (Unimodal functions)

| Functions | | PSO | FA | CSO | GWO | SCA | MPA | AOA | DSA |
|---|---|---|---|---|---|---|---|---|---|
| F1 | Best | 1.09E + 02 | 2.03E−01 | 7.72E−09 | 1.18E−09 | 1.90E+01 | 1.30E−07 | 9.71E−45 | **2.95E−133** |
| | Mean | 6.63E + 02 | 2.87E−01 | 4.84E−05 | 8.95E−09 | 7.19E+02 | 5.10E−07 | 1.57E−33 | **2.33E−100** |
| | Std | 3.98E+02 | 4.34E−02 | 1.07E−04 | 9.09E−09 | 7.88E+02 | 4.03E−07 | 6.06E−33 | **1.18E−99** |
| | Time/s | **1.59E−02** | 1.29E−01 | 4.09E−02 | 3.62E−02 | 2.89E−02 | 7.49E−02 | 3.33E−02 | 4.99E−02 |
| F2 | Best | 1.01E−01 | 7.19E−01 | 5.39E−09 | 1.53E−06 | 2.34E−01 | 2.34E−05 | 1.20E−25 | **1.39E−65** |
| | Mean | 1.91E−01 | 1.12E+00 | 6.51E−07 | 6.08E−06 | 1.13E+00 | 1.18E−04 | 4.33E−18 | **9.42E−51** |
| | Std | 4.99E−02 | 1.39E−01 | 9.75E−07 | 3.16E-06 | 1.24E+00 | 6.14E−05 | 2.20E−17 | **5.09E−50** |
| | Time/s | **1.60E−02** | 1.24E−01 | 4.27E−02 | 4.11E−02 | 2.94E−02 | 7.00E−02 | 3.57E−02 | 4.85E−02 |
| F3 | Best | 1.31E+03 | 8.06E+02 | 2.41E+03 | 3.19E−01 | 7.05E+03 | 5.79E−01 | 8.66E−38 | **1.31E−129** |
| | Mean | 8.98E+03 | 1.65E+03 | 7.19E+03 | 5.96E+00 | 1.95E+04 | 6.61E+00 | 1.28E−25 | **1.32E−89** |
| | Std | 4.90E+03 | 6.43E+02 | 2.68E+03 | 6.44E+00 | 7.45E+03 | 7.10E+00 | 4.47E−25 | **7.26E−89** |
| | Time/s | **6.63E−02** | 1.74E−01 | 9.53E−02 | 8.62E−02 | 7.90E−02 | 1.83E−01 | 8.39E−02 | 1.48E−01 |
| F4 | Best | 3.95E+00 | 2.04E−01 | 1.62E+01 | 9.82E−03 | 2.04E+01 | 1.99E−03 | 1.66E−23 | **7.06E−68** |
| | Mean | 6.51E+00 | 2.80E−01 | 3.07E+01 | 3.50E−02 | 5.75E+01 | 4.24E−03 | 3.39E−16 | **2.47E−48** |
| | Std | 1.37E+00 | 6.49E−02 | 6.98E+00 | 1.80E−02 | 1.22E+01 | 1.26E−03 | 1.12E−15 | **1.32E−47** |
| | Time/s | **1.61E−02** | 1.24E−01 | 4.24E−02 | 3.64E−02 | 2.90E−02 | 6.88E−02 | 3.61E−02 | 4.53E−02 |
| F5 | Best | **1.98E+01** | 3.58E+01 | 2.80E+01 | 2.62E+01 | 2.93E+04 | 2.63E+01 | 2.86E+01 | 2.89E+01 |
| | Mean | 7.78E+01 | 1.19E+02 | 8.73E+03 | **2.77E+01** | 1.30E+06 | 2.71E+01 | 2.89E+01 | 2.89E+01 |
| | Std | 7.98E+01 | 1.32E+02 | 4.40E+04 | 7.84E−01 | 1.29E+06 | 5.15E−01 | 7.14E−02 | **2.92E−02** |
| | Time/s | **2.30E−02** | 1.35E−01 | 5.13E−02 | 4.20E−02 | 3.52E−02 | 9.13E−02 | 4.01E−02 | 6.03E−02 |
| F6 | Best | 7.24E−28 | 4.20E−02 | 3.07E−72 | 1.27E−99 | 9.86E−31 | 1.04E−41 | 5.48E−104 | **2.46E−153** |
| | Mean | 2.09E−22 | 3.13E+02 | 1.99E−58 | 2.43E−79 | 5.49E−23 | 2.93E−23 | 9.08E−72 | **4.09E−119** |
| | Std | 1.08E−21 | 4.49E+02 | 1.03E−57 | 1.28E−78 | 3.00E−22 | 1.59E−22 | 4.97E−71 | **2.24E−118** |
| | Time/s | **1.70E−02** | 1.30E−01 | 4.37E−02 | 3.64E−02 | 2.95E−02 | 7.52E−02 | 3.47E−02 | 4.55E−02 |
| F7 | Best | 7.77E−02 | 8.94E−03 | 1.14E−02 | 1.29E−03 | 9.42E−02 | 1.27E−03 | 2.91E−05 | **1.61E−05** |
| | Mean | 2.17E−01 | 5.10E−02 | 1.96E−01 | 5.77E−03 | 1.54E+00 | 4.23E−03 | 1.71E−03 | **4.33E−04** |
| | Std | 7.70E−02 | 3.82E−02 | 2.68E−01 | 2.78E−03 | 1.92E+00 | 1.98E−03 | 1.08E−03 | **5.28E−04** |
| | Time/s | **4.09E−02** | 1.54E−01 | 6.67E−02 | 6.14E−02 | 5.35E−02 | 1.29E−01 | 5.84E−02 | 9.44E−02 |

Bold values indicate optimal value of the comparison algorithms in this study exploration and exploitation percentage curves of the population diversity during the process of optima with 200 iterations and boxplot of the comparison algorithms for the benchmark functions, respectively. Eventually, DSA successfully produces effective results that verify its performance as we will illustrate in this section.

## 5.1 Statistical results analysis

The statistical results are reported in Tables 3, 4 and 5, respectively. Table 3 shows that DSA displayed an extremely good exploitation ability among the comparison algorithms except F5. According to Table 4, the DSA yields a pretty exploration ability for multimodal dimension problems, excluding F12 and F13. For functions F9 and F10, AOA and DSA can obtain the best fitness value, but the Mean and Std of AOA are much worse than DSA. For F14, CSO, MPA, AOA and DSA can obtain the best optimum. According to the dimension of benchmark functions, it can be divided into three types: low dimension (Less than 10 dimension), high dimension (Between 10 and 300 dimension), and large-scale (Greater than 300 dimension). In general, DSA demonstrates outstanding performance on test functions F1, F2, F3, F4, F6, F7, F8, F9, F10, F11, and F14, especially on F9 and F11 because of the best fitness obtained by the proposed DSA.

In addition, from the Best in Table 5, DSA can obtain the best fitness on functions F15, F16, F17, and F18. It illustrates the advantages of the DSA to strike a balance between exploration and exploitation phases for fixed dimension problems. However, for the Mean and Std on fixed functions, MPA has better stability on F15 and F16 than DSA. PSO algorithm has a best stability on F18 among the comparison algorithms. Thus, the stability of the DSA on fixed dimension functions should be improved in further study.

**Table 4** Comparison statistical results (Multimodal functions)

| Functions | | PSO | FA | CSO | GWO | SCA | MPA | AOA | DSA |
|---|---|---|---|---|---|---|---|---|---|
| F8 | Best | − 3.86E+03 | − 7.06E+03 | − 7.62E+03 | − 8.01E+03 | − 4.31E+03 | − 9.11E+03 | − 1.04E+08 | **− 9.43E+03** |
| | Mean | − 2.68E+03 | − 5.50E+03 | − 6.40E+03 | − 5.80E+03 | − 3.50E+03 | − 7.98E+03 | − 4.86E+06 | **− 5.77E+03** |
| | Std | 3.90E+02 | 7.57E+02 | 6.32E+02 | 1.30E+03 | **2.81E+02** | 5.32E+02 | 2.02E+07 | 1.38E+03 |
| | Time/s | **2.46E−02** | 1.34E−01 | 5.36E−02 | 4.48E−02 | 3.64E−02 | 9.38E−02 | 4.12E−02 | 6.22E−02 |
| F9 | Best | 2.69E+01 | 2.04E+01 | 3.22E−09 | 1.03E+00 | 9.42E+00 | 9.67E−07 | **0** | **0** |
| | Mean | 4.88E+01 | 3.29E+01 | 3.69E+00 | 1.33E+01 | 7.07E+01 | 1.54E−03 | 2.17E+01 | **0** |
| | Std | 1.00E+01 | 9.35E+00 | 1.34E+01 | 5.99E+00 | 4.08E+01 | 5.65E−03 | 5.64E+01 | **0** |
| | Time/s | **2.14E−02** | 1.32E−01 | 4.43E−02 | 3.92E−02 | 3.34E−02 | 7.59E−02 | 3.59E−02 | 5.07E−02 |
| F10 | Best | 3.46E−02 | 3.13E−01 | 3.75E−05 | 6.60E−06 | 2.77E+00 | 3.76E−05 | 7.99E−15 | **8.88E−16** |
| | Mean | 1.57E−01 | 4.62E−01 | 8.60E−04 | 1.89E−05 | 1.53E+01 | 1.42E−04 | 1.66E+01 | **8.88E−16** |
| | Std | 2.53E−01 | 6.48E−02 | 2.16E−03 | 1.11E−05 | 7.19E+00 | 5.48E−05 | 7.57E+00 | **0** |
| | Time/s | **2.14E−02** | 1.31E−01 | 4.67E−02 | 3.95E−02 | 3.54E−02 | 7.59E−02 | 3.97E−02 | 5.16E−02 |
| F11 | Best | 2.80E+02 | 4.48E−01 | 4.63E−08 | 2.11E−09 | 1.21E+00 | 7.69E−08 | **0** | **0** |
| | Mean | 3.14E+02 | 5.76E−01 | 4.77E−02 | 1.36E−02 | 6.99E+00 | 3.65E−06 | 1.38E−02 | **0** |
| | Std | 2.42E+01 | 6.41E−02 | 1.22E−01 | 1.58E−02 | 5.84E+00 | 5.86E−06 | 5.32E−02 | **0** |
| | Time/s | **2.91E−02** | 1.33E−01 | 5.34E−02 | 4.63E−02 | 4.00E−02 | 8.76E−02 | 4.30E−02 | 6.31E−02 |
| F12 | Best | 4.08E−03 | **1.66E−03** | 2.50E−01 | 3.45E−02 | 1.14E+01 | 1.90E−03 | 5.64E−01 | 2.93E−01 |
| | Mean | 1.86E+00 | **4.95E−03** | 6.86E+04 | 1.09E−01 | 3.07E+06 | 1.06E−02 | 8.09E−01 | 7.52E−01 |
| | Std | 1.09E+00 | **3.89E−03** | 3.20E+05 | 5.29E−02 | 9.02E+06 | 7.35E−03 | 1.39E−01 | 3.39E−01 |
| | Time/s | **8.97E−02** | 1.94E−01 | 1.19E−01 | 1.05E−01 | 1.01E−01 | 2.24E−01 | 1.07E−01 | 1.86E−01 |
| F13 | Best | **8.37E−04** | 2.24E−02 | 1.71E+00 | 2.38E−01 | 2.41E+01 | 3.44E−02 | 2.51E+00 | 1.75E+00 |
| | Mean | **3.24E−02** | 3.14E−02 | 1.00E+05 | 1.08E+00 | 1.00E+07 | 2.51E−01 | 2.92E+00 | 2.84E+00 |
| | Std | **8.68E−02** | 6.57E−03 | 2.68E+05 | 3.32E−01 | 1.78E+07 | 1.34E−01 | 8.78E−02 | 3.26E−01 |
| | Time/s | **9.43E−02** | 2.00E−01 | 1.21E−01 | 1.08E−01 | 1.00E−01 | 2.31E−01 | 1.04E−01 | 1.83E−01 |
| F14 | Best | **0** | 13.3017211 | **0** | 4.27E+00 | 4.80E+00 | **0** | **0** | **0** |
| | Mean | 3.65E+00 | 1.72E+01 | **0** | 8.57E+00 | 1.06E+01 | **0** | **0** | **0** |
| | Std | 3.08E+00 | 1.87+00 | **0** | 2.60E+00 | 1.92E+00 | **0** | **0** | **0** |
| | Time/s | 9.51E−01 | 1.00E+00 | 9.97E−01 | 9.43E−01 | **9.41E−01** | 1.96E+00 | 9.45E−01 | 1.94E+00 |

Bold values indicate optimal value of the comparison algorithms in this study

Moreover, DSA is compared with the other seven algorithms in the running-time calculation on the eighteen benchmark functions. The running-time calculation method is that the comparison methods independently run 30 times on each test function and noted the results in Table 3, 4, 5, respectively. DSA outperforms FA and MPA while taking less time than for unimodal, multimodal, and fixed test functions. Compared with the running-time of the CSO, GWO, and AOA, the running-time of the DSA is an order of magnitude, and the gap is small. Although the running-time of the PSO algorithm is the shortest, and SCA followed, their optimization accuracy is poor among the comparison algorithms. Generally, the proposed DSA still possesses effective superiorities over the comparison methods on the running-time.

Two of the frequently used tests are used to statistically evaluate the performance of the DSA in this paper. Tables 6 and 7 illustrate Friedman rank and Wilcoxon's rank-sum test results. According to the Friedman test results listed in Table 6, it can be concluded that the rankings of the eight comparison algorithms are DSA > MPA > AOA > GWO > CSO > PSO > FA > SCA. It is shown that DSA can produce satisfactory results and is also statistically superior to comparison algorithms. DSA will play a constructive role in the future as a robust algorithm.

The WSR results of pair-wise comparison of the DSA and comparison methods are presented in Table 7 at 0.05 significance level. Where H = 1 means acceptable; H = 0 means rejection; and NaN means that the optimization values of the two algorithms are similar. According to the p-values in Table 7, for function F8, the FA, GWO, AOA are better than DSA. For functions F11 and F12, AOA is better than DSA from the WSR test. For function F14, the optimum values of CSO, MPA, and AOA are similar to DSA. Additionally, PSO is better than DSA on functions F15 and F16 based on WSR test results. Also, the WSR test

**Table 5** Comparison statistical results (fixed functions)

| Functions | | PSO | FA | CSO | GWO | SCA | MPA | AOA | DSA |
|---|---|---|---|---|---|---|---|---|---|
| F15 | Best | **9.98E−01** | 1.00E+00 | **9.98E−01** | **9.98E−01** | **9.98E−01** | **9.98E−01** | **9.98E−01** | **9.98E−01** |
| | Mean | 1.89E+00 | 5.35E+00 | 1.75E+00 | 5.14E+00 | 2.82E+00 | **9.98E−01** | 1.26E+00 | 1.95E+00 |
| | Std | 1.78E+00 | 2.65E+00 | 2.18E+00 | 4.24E+00 | 2.85E+00 | **5.17E−16** | 6.92E−01 | 2.89E−04 |
| | Time/s | **1.35E−01** | 2.20E−01 | 1.69E−01 | 1.37E−01 | 1.39E−01 | 3.04E−01 | 1.46E−01 | 2.94E−01 |
| F16 | Best | 3.14E−04 | 3.27E−04 | 4.32E−04 | 3.20E−04 | 4.33E−04 | **3.07E−04** | 3.36E−04 | **3.07E−04** |
| | Mean | 6.19E−04 | 2.29E−03 | 7.85E−04 | 4.56E−03 | 1.15E−03 | **3.13E−04** | 1.21E−03 | 9.80E−04 |
| | Std | 3.59E−04 | 7.09E−03 | 2.49E−04 | 8.04E−03 | 3.85E−04 | **1.56E−05** | 1.19E−03 | 7.40E−04 |
| | Time/s | **1.07E−02** | 1.01E−01 | 3.97E−02 | 1.52E−02 | 1.48E−02 | 5.67E−02 | 1.90E−02 | 5.00E−02 |
| F17 | Best | **− 1.0316** | **− 1.0316** | **− 1.0316** | **− 1.0316** | **− 1.0316** | **− 1.0316** | **− 1.0316** | **− 1.0316** |
| | Mean | **− 1.0316** | **− 1.0316** | **− 1.0316** | **− 1.0316** | − 1.0315 | **− 1.0316** | − 1.0311 | **− 1.0316** |
| | Std | 5.53E−16 | 6.78E−05 | 8.22E−08 | 1.82E−07 | 2.20E−04 | 5.66E−14 | 6.69E−04 | **0.00E+00** |
| | Time/s | **7.43E−03** | 9.55E−02 | 3.63E−02 | 1.26E−02 | 1.25E−02 | 4.99E−02 | 1.72E−02 | 4.37E−02 |
| F18 | Best | **0.398** | **0.398** | **0.398** | **0.398** | **0.398** | **0.398** | **0.398** | **0.398** |
| | Mean | **0.398** | **0.398** | **0.398** | **0.398** | 0.407 | **0.398** | 0.435 | **0.398** |
| | Std | **0.00E+00** | 3.64E−06 | 8.87E−05 | 2.14E−04 | 1.71E−02 | 2.49E−12 | 5.37E−02 | 9.66E−05 |
| | Time/s | **5.35E−03** | 9.73E−02 | 3.50E−02 | 1.05E−02 | 9.98E−03 | 4.51E−02 | 1.44E−02 | 4.02E−02 |

Bold values indicate optimal value of the comparison algorithms in this study

**Table 6** Comparison results of Friedman rank test

| Functions | PSO | FA | CSO | GWO | SCA | MPA | AOA | DSA |
|---|---|---|---|---|---|---|---|---|
| F1 | 7.57 | 6.00 | 4.63 | 3.00 | 7.43 | 4.37 | 2.00 | 1.00 |
| F2 | 6.03 | 7.77 | 3.03 | 3.97 | 7.20 | 5.00 | 2.00 | 1.00 |
| F3 | 6.80 | 5.03 | 6.40 | 3.47 | 7.77 | 3.53 | 2.00 | 1.00 |
| F4 | 6.00 | 5.00 | 7.03 | 4.00 | 7.97 | 3.00 | 2.00 | 1.00 |
| F5 | 5.50 | 6.47 | 4.93 | 1.97 | 7.97 | 1.33 | 3.50 | 4.33 |
| F6 | 6.87 | 8.00 | 4.00 | 2.47 | 5.67 | 5.47 | 2.53 | 1.00 |
| F7 | 6.83 | 5.20 | 6.13 | 3.77 | 7.80 | 3.17 | 1.97 | 1.13 |
| F8 | 7.90 | 4.47 | 3.33 | 3.90 | 6.80 | 1.37 | 4.27 | 3.97 |
| F9 | 7.13 | 6.10 | 3.23 | 4.87 | 7.23 | 3.63 | 2.38 | 1.42 |
| F10 | 5.23 | 6.10 | 4.00 | 2.17 | 7.67 | 3.33 | 6.50 | 1.00 |
| F11 | 8.00 | 6.00 | 4.37 | 3.80 | 7.00 | 3.63 | 1.73 | 1.47 |
| F12 | 5.90 | 1.17 | 6.50 | 3.03 | 7.97 | 1.87 | 4.83 | 4.73 |
| F13 | 1.10 | 1.90 | 6.30 | 4.00 | 7.97 | 3.00 | 5.60 | 6.13 |
| F14 | 5.10 | 8.00 | 2.52 | 6.10 | 6.73 | 2.52 | 2.52 | 2.52 |
| F15 | 2.97 | 7.23 | 3.63 | 6.53 | 6.00 | 1.70 | 4.77 | 3.17 |
| F16 | 3.47 | 5.43 | 4.93 | 4.57 | 6.43 | 1.07 | 5.43 | 4.67 |
| F17 | 1.47 | 6.13 | 1.87 | 4.97 | 7.30 | 2.80 | 7.57 | 3.90 |
| F18 | 1.33 | 5.10 | 2.53 | 5.67 | 7.30 | 2.67 | 7.63 | 3.77 |
| Total | 95.20 | 101.10 | 79.36 | 72.26 | 130.21 | 53.46 | 69.23 | **47.21** |
| Avg | 5.29 | 5.62 | 4.41 | 4.01 | 7.23 | 2.97 | 3.85 | **2.62** |
| Rank | 6 | 7 | 5 | 4 | 8 | 2 | 3 | **1** |

Bold values indicate optimal value of the comparison algorithms in this study results of FA, CSO, and AOA is better than DSA with 200 iterations on F15 and F16.

Overall, the statistical results verify that there is a significant difference between the results obtained by the DSA and the comparison approaches in almost all cases. Especially, DSA in benchmark functions F1–F4, F6, F7, F9, F10, F11, F14, F17 and F18 has a significant advantage over almost all comparison methods. However, in functions

**Table 7** Comparison results of Wilcoxon rank-sum test

| Functions | PSO vs. DSA | | | FA vs. DSA | | | CSO vs. DSA | | |
|---|---|---|---|---|---|---|---|---|---|
| | p-value | H | Z-value | p-value | H | Z-value | p-value | H | Z-value |
| F1 | 3.02E−11 | 1 | 6.65 | 3.02E−11 | 1 | 6.65 | 3.02E−11 | 1 | 6.65 |
| F2 | 3.02E−11 | 1 | 6.65 | 3.02E−11 | 1 | 6.65 | 3.02E−11 | 1 | 6.65 |
| F3 | 3.02E−11 | 1 | 6.65 | 3.02E−11 | 1 | 6.65 | 3.02E−11 | 1 | 6.65 |
| F4 | 3.02E−11 | 1 | 6.65 | 3.02E−11 | 1 | 6.65 | 3.02E−11 | 1 | 6.65 |
| F5 | 1.11E−06 | 1 | 4.87 | 3.02E−11 | 1 | 6.65 | 1.33E−02 | 1 | 2.48 |
| F6 | 3.02E−11 | 1 | 6.65 | 3.02E−11 | 1 | 6.65 | 3.02E−11 | 1 | 6.65 |
| F7 | 3.02E−11 | 1 | 6.65 | 3.02E−11 | 1 | 6.65 | 3.02E−11 | 1 | 6.65 |
| F8 | 3.02E−11 | 1 | 6.65 | 8.88E−01 | 0 | 0.14 | 5.83E−03 | 1 | −2.76 |
| F9 | 1.21E−12 | 1 | 7.10 | 1.21E−12 | 1 | 7.10 | 1.21E−12 | 1 | 7.10 |
| F10 | 1.21E−12 | 1 | 7.10 | 1.21E−12 | 1 | 7.10 | 1.21E−12 | 1 | 7.10 |
| F11 | 1.21E−12 | 1 | 7.10 | 1.21E−12 | 1 | 7.10 | 1.21E−12 | 1 | 7.10 |
| F12 | 5.09E−06 | 1 | 4.56 | 3.02E−11 | 1 | − 6.65 | 5.09E−06 | 1 | 4.56 |
| F13 | 3.02E−11 | 1 | − 6.65 | 3.02E−11 | 1 | − 6.65 | 8.24E−02 | 0 | 1.74 |
| F14 | 4.57E−12 | 1 | 6.92 | 1.21E−12 | 1 | 7.10 | NaN | 0 | NaN |
| F15 | 3.18E−01 | 0 | − 1.00 | 1.72E−12 | 1 | 7.06 | 7.24E−01 | 0 | 0.35 |
| F16 | 1.58E−01 | 0 | − 1.41 | 2.40E−01 | 0 | 1.18 | 8.07E−01 | 0 | 0.24 |
| F17 | 4.17E−13 | 1 | − 7.25 | 1.21E−12 | 1 | 7.10 | 9.43E−09 | 1 | −5.74 |
| F18 | 4.16E−14 | 1 | − 7.56 | 1.12E−10 | 1 | 6.45 | 2.17E−03 | 1 | −3.07 |
| NaN/0/1 | 0/2/16 | | | 0/2/16 | | | 1/3/14 | | |

| Functions | GWO vs. DSA | | | SCA vs. DSA | | | MPA vs. DSA | | | AOA vs. DSA | | |
|---|---|---|---|---|---|---|---|---|---|---|---|---|
| | p-value | H | Z-value | p-value | H- | Z-value | p-value | H | Z-value | p-value | H | Z-value |
| F1 | 3.02E−11 | 1 | 6.65 | 3.02E−11 | 1 | 6.65 | 3.02E−11 | 1 | 6.65 | 3.02E−11 | 1 | 6.65 |
| F2 | 3.02E−11 | 1 | 6.65 | 3.02E−11 | 1 | 6.65 | 3.02E−11 | 1 | 6.65 | 3.02E−11 | 1 | 6.65 |
| F3 | 3.02E−11 | 1 | 6.65 | 3.02E−11 | 1 | 6.65 | 3.02E−11 | 1 | 6.65 | 3.02E−11 | 1 | 6.65 |
| F4 | 3.02E−11 | 1 | 6.65 | 3.02E−11 | 1 | 6.65 | 3.02E−11 | 1 | 6.65 | 3.02E−11 | 1 | 6.65 |
| F5 | 3.02E−11 | 1 | −6.65 | 3.02E−11 | 1 | 6.65 | 3.02E−11 | 1 | − 6.65 | 8.88E−06 | 1 | −4.44 |
| F6 | 3.02E−11 | 1 | 6.65 | 3.02E−11 | 1 | 6.65 | 3.02E−11 | 1 | 6.65 | 3.02E−11 | 1 | 6.65 |
| F7 | 4.50E11 | 1 | 6.59 | 3.02E−11 | 1 | 6.65 | 6.70E−11 | 1 | 6.53 | 8.20E−07 | 1 | 4.93 |
| F8 | 3.04E−01 | 0 | − 1.03 | 3.34E−11 | 1 | 6.63 | 4.31E−08 | 1 | − 5.48 | 4.29E−01 | 0 | 0.79 |
| F9 | 1.21E−12 | 1 | 7.10 | 1.21E−12 | 1 | 7.10 | 1.21E−12 | 1 | 7.10 | 2.16E−02 | 1 | 2.30 |
| F10 | 1.21E−12 | 1 | 7.10 | 1.21E−12 | 1 | 7.10 | 1.21E−12 | 1 | 7.10 | 6.11E−13 | 1 | 7.20 |
| F11 | 1.21E−12 | 1 | 7.10 | 1.21E−12 | 1 | 7.10 | 1.21E−12 | 1 | 7.10 | 1.61E−01 | 0 | 1.40 |
| F12 | 3.02E−11 | 1 | − 6.65 | 3.02E−11 | 1 | 6.65 | 3.02E−11 | 1 | − 6.65 | 4.04E−01 | 0 | 0.84 |
| F13 | 3.02E−11 | 1 | − 6.65 | 3.02E−11 | 1 | 6.65 | 3.02E−11 | 1 | − 6.65 | 1.11E−03 | 1 | −3.26 |
| F14 | 1.21E−12 | 1 | 7.10 | 1.21E−12 | 1 | 7.10 | NaN | 0 | NaN | NaN | 0 | NaN |
| F15 | 6.95E−11 | 1 | 6.52 | 3.77E−12 | 1 | 6.95 | 1.07E−12 | 1 | − 7.12 | 7.29E−09 | 1 | 5.78 |
| F16 | 6.63E−01 | 0 | 0.44 | 4.84E−02 | 1 | 1.97 | 4.57E−09 | 1 | − 5.86 | 2.97E−01 | 0 | 1.04 |
| F17 | 1.21E−12 | 1 | 7.10 | 1.21E−12 | 1 | 7.10 | 1.21E−12 | 1 | − 7.10 | 1.21E−12 | 1 | 7.10 |
| F18 | 6.03E−11 | 1 | 6.54 | 2.96E−12 | 1 | 6.98 | 2.37E−12 | 1 | − 7.01 | 3.69E−12 | 1 | 6.95 |
| NaN/0/1 | 0/1/17 | | | 0/0/18 | | | 1/0/17 | | | 1/4/13 | | |

F15 and F16, the performance of the DSA in the comparison algorithms is insufficient, which indicates that the DSA needs further improvement for solving fixed-dimensional problems. This conclusion of the statistical tests is in line.

## 5.2 Convergence curves analysis

Notably, to completely report the performance of the competitive algorithms, the convergence curves of the eighteen test functions are shown in Fig. 7. The y-axis and x-axis represent the fitness value and iteration, respectively. The convergence curves show that the poor performance of the SCA, PSO and FA is the imbalance between the exploitation and exploration phases. The convergence curve clearly illustrates the advantages of the DSA integrating the two phases into the search process. Except for DSA, MPA and AOA, almost all comparison algorithms converge to a local optimum for test functions. It also indicates that the DSA is faster and superior to other comparison algorithms in solving almost all the numerical optimization problems, except F5, F12, F13, and F17.

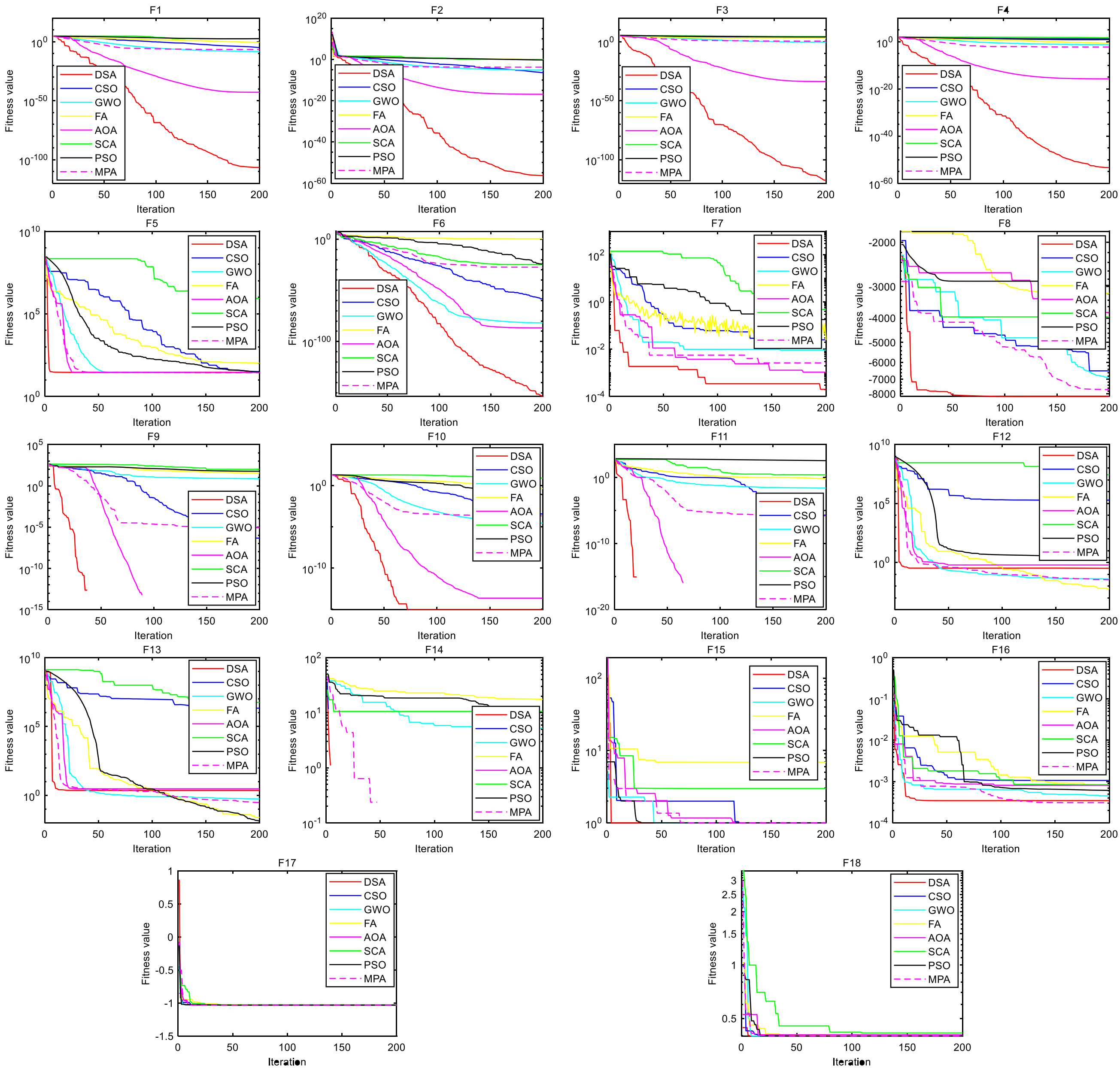

**Fig. 7** Convergence curves of eighteen benchmark functions with 30 times

## 5.3 Diversity analysis

In this study, the components DSA exploration and development capabilities of the impact of the diversity were analyzed. The plots are discussed to evaluate the ability of the DSA to balance exploration and exploitation. Figure 8 shows the exploration and exploitation percentage curves of population diversity in the search space while solving the test functions.

As shown in Fig. 8, DSA preserves a balance between the exploration and exploitation rates during the search process for all agents in solving the benchmark optimization functions. Notably, the results balance the capabilities between exploration and exploitation phases to push duck $i$ to the global optimal solution by moving to the optimum produced at a given time.

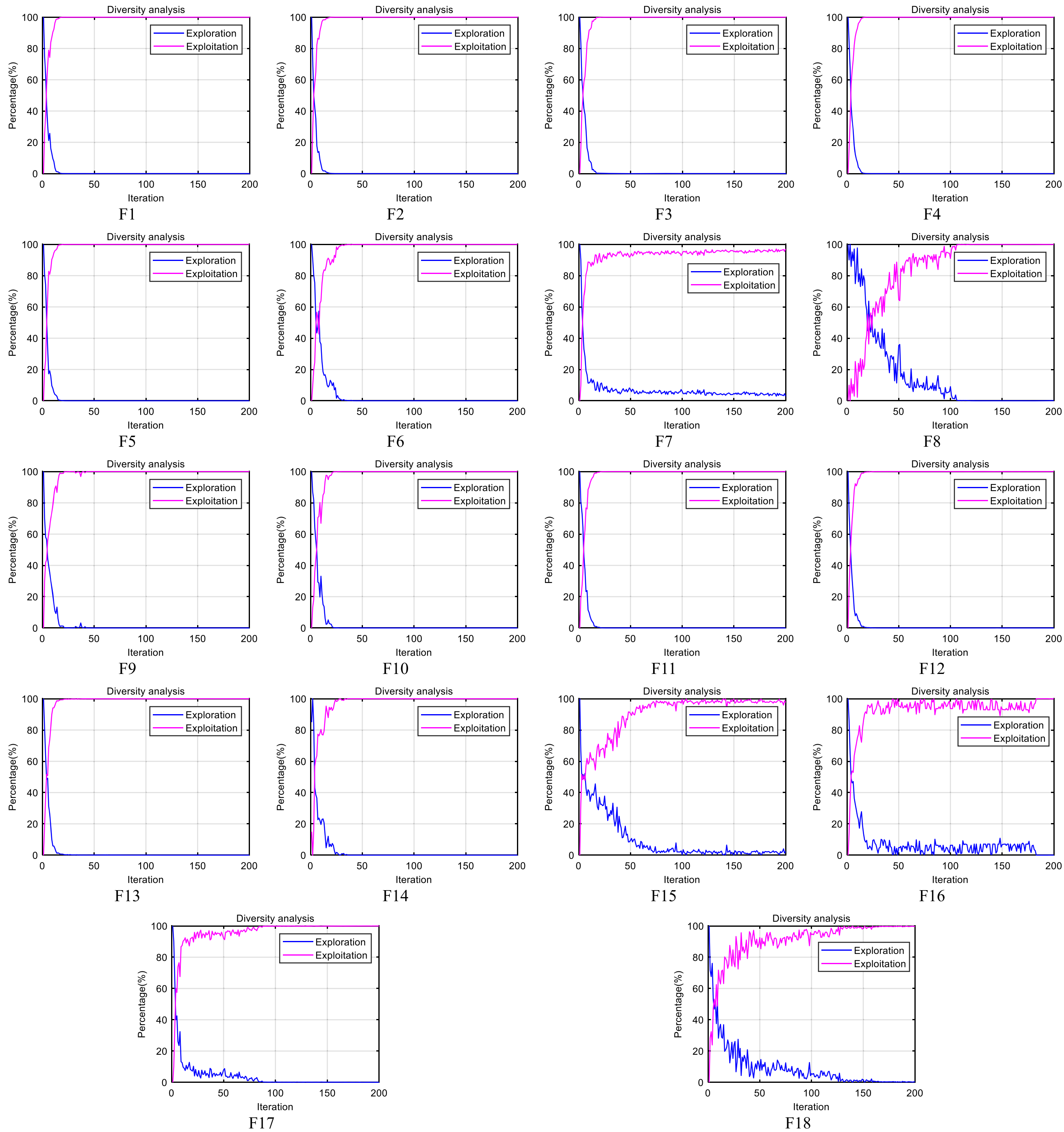

**Fig. 8** Average diversity analysis of eighteen benchmark functions with 30 times

According to Fig. 8 and Table 5, although DSA can obtain the best fitness value, the Mean and Std of MPA on F15 and F16 are better than DSA. In other words, the exploration and exploitation ability of the DSA should be improved for the fixed-dimension or low-dimensional optimization problems.

## 5.4 Boxplot analysis

The boxplots of the comparison algorithms on the test functions (F1–F16) are illustrated in Fig. 9. According to Tables 6 and 7 and the charts shown in Fig. 9, it can be ascertained that the DSA achieved the best results and had the best convergence among the others on most of the functions. However, the FA performs better than DSA in F12, and the PSO algorithm performs better than DSA in F13, respectively. Consequently, it can be inferred that the DSA outperforms other comparison methods on classical benchmark functions.

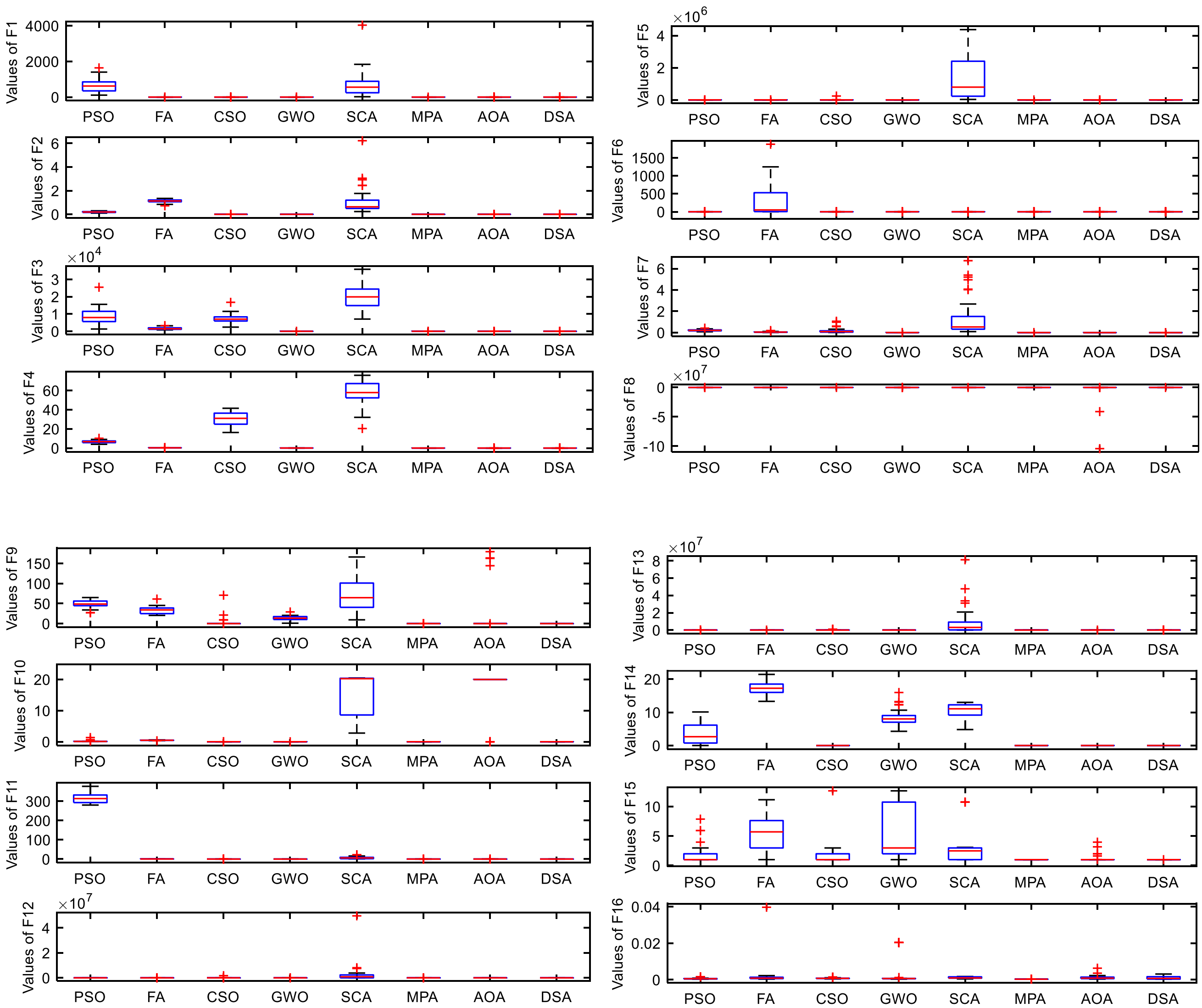

**Fig. 9** Boxplot of functions F1-F16

## 5.5 Sensitivity analysis of the Scalability test

In this subsection, parameter sensitivity tests are used to assess the influence of population size, and iterations on the proposed method that four test functions (two unimodal and two multimodal) are selected, including Sphere function, Schwefel 2.22 function, Rastrigin function, and Ackley function. The population size is set to 30, 50, 80 and 100, and the number of iterations is set to 200, 500, 1000 and 2000, with the dimension of test functions is fixed to 30.

The results of the sensitivity analysis (the mean fitness, Std and convergence curves) for the above four functions are illustrated as follows: (i) Number of ducks (*N*), (ii) Max iterations (*T*). DSA was simulated for different numbers of ducks with iteration is fixed to 500. Table 8 shows the Mean and Std value of the DSA when it was applied to solve Sphere (F1), Schwefel 2.22 (F2), Rastrigin (F9), and Ackley (F10) with different number of ducks.

Figure 10 shows the convergence curves of the DSA on the four functions related to number of ducks, respectively. DSA is simulated for various numbers of iterations. Table 9 displays the Mean and Std value of the DSA when it is applied to simulate four test functions with different numbers of iterations. Figure 10 plots the convergence curves of the DSA for Sphere (F1), Schwefel 2.22 (F2),

**Table 8** Comparison results of DSA using different values for ducks with 500 iterations

| Function | Dim = 30 | $N$ = 30 | $N$ = 50 | $N$ = 80 | $N$ = 100 |
|---|---|---|---|---|---|
| F1 | Mean | 5.08E−262 | **6.69E−264** | 2.90E−260 | 3.94E−255 |
| | Std | 0.00E+00 | **0.00E+00** | 0.00E+00 | 0.00E+00 |
| F2 | Mean | **1.74E−139** | 8.44E−123 | 5.70E−127 | 1.71E−130 |
| | Std | **7.27E−139** | 4.62E−122 | 2.59E−126 | 9.36E130 |
| F9 | Mean | 0.00E+00 | 0.00E+00 | 0.00E+00 | 0.00E+00 |
| | Std | 0.00E+00 | 0.00E+00 | 0.00E+00 | 0.00E+00 |
| F10 | Mean | 8.88E−16 | 8.88E−16 | 8.88E−16 | 8.88E−16 |
| | Std | 0.00E+00 | 0.00E+00 | 0.00E+00 | 0.00E+00 |

Bold values indicate optimal value of the comparison algorithms in this study

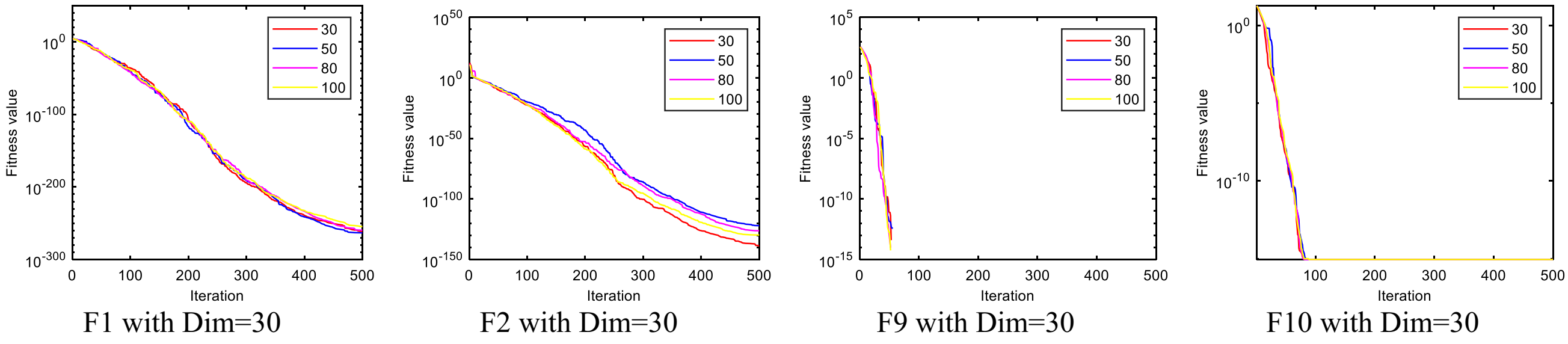

**Fig. 10** Sensitivity analysis of the proposed DSA for number of search agents

Rastrigin (F9), and Ackley (F10) using different numbers of iterations.

Table 8 shows that the best performance is obtained with different search agents. As can be seen visually from Fig. 9 and Table 9, as the population size increased, the Mean and Std become slightly worse, except functions F9 and F10. The reason is that the search ability of the DSA has hardly changed after the population reaches 30. As can be seen from Fig. 10, the sensitivity of the proposed DSA increases slightly with the number of search agents.

The results in Table 9 and Fig. 11 prove that DSA converges to the optimum when the number of iterations increases. This supports the importance of the iteration number on the robustness and convergence behavior of the DSA. As can be seen visually from Fig. 10 and Table 9, as the population size increased, the Mean and Std become better, except functions F9 and F10. The reason is that the increase in iterations increases the number of searches and accuracy. However, the mean fitness and Std of the F9 and F10 do not become better as the number of iterations increased.

According to the above analysis, when the global approximate optimal solution is roughly found, as the population and the number of iterations continue to grow, the result does not increase proportionally. The results are not increased proportionally. In a word, researchers can set the favorable population and number of iterations according to specific tasks.

## 6 Applications of the proposed DSA

To confirm the performance of the DSA for solving real-world engineering constrained optimization problems are presented: Three-bar truss problem (TBTP) [56], and Sawmill operation problem (SOP) [57]. Other four engineering constrained optimization problems are also present in subSect. 6.3. SubSect. 6.4 gives the comparison results of the designed DSA for solving the node optimization deployment of the WSN. All the considered problems have several inequality constraints that should be handled.

### 6.1 Three-bar truss problem

The schematic of Three-bar truss (TBS) structure is shown in Fig. 12. The volume of a statically loaded Three-bar truss is to be minimized subject to stress ($\sigma$) constraints on each of the truss members. The objective is to assess the optimal cross-sectional areas ($A_1$, $A_2$). This problem can be expressed as below (volume of a member = cross-sectional area × length):

$$\min : f(x_1, x_2) = f(A_1, A_2) = (2\sqrt{2}A_1 + A_2) \cdot l \qquad (17)$$

**Table 9** Comparison results of the DSA with different iterations

| Function | Dim = 30 | $T$ = 200 | $T$ = 500 | $T$ = 1000 | $T$ = 2000 |
|---|---|---|---|---|---|
| F1 | Mean | 1.60E−84 | 1.23E−272 | 0 | **0** |
| | Std | 8.78E−84 | 0 | 0 | **0** |
| F2 | Mean | 8.25E−50 | 1.74E−130 | 1.18E−279 | **0** |
| | Std | 3.21E−49 | 9.51E−130 | 0 | **0** |
| F9 | Mean | 0 | 0 | 0 | **0** |
| | Std | 0 | 0 | 0 | **0** |
| F10 | Mean | 8.88E−16 | 8.88E−16 | 8.88E−16 | **8.88E−16** |
| | Std | 0 | 0 | 0 | **0** |

Bold values indicate optimal value of the comparison algorithms in this study

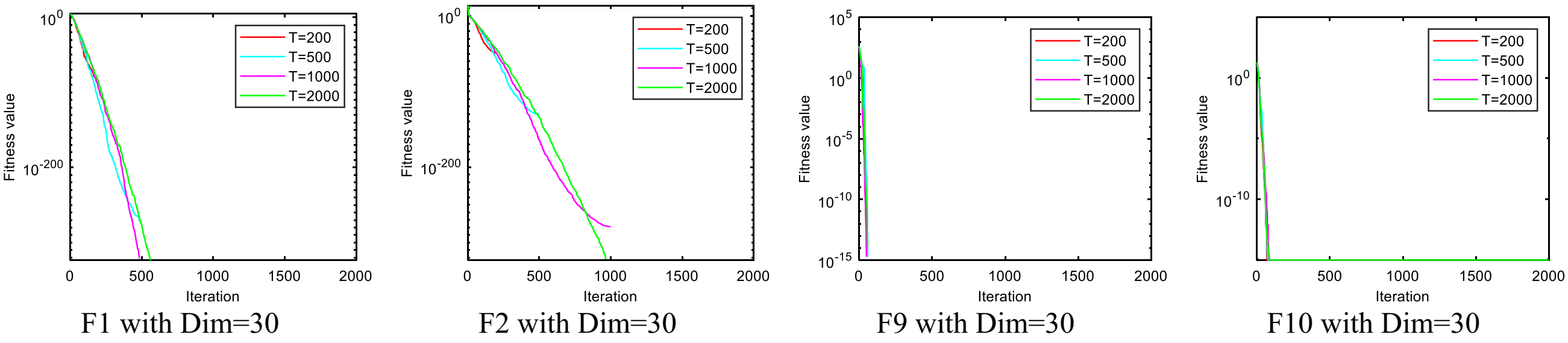

**Fig. 11** Sensitivity analysis of the proposed DSA for the number of iterations

$$\text{s.t}\begin{cases} g_1 = \dfrac{\sqrt{2}A_1 + A_2}{\sqrt{2}A_1^2 + 2A_1A_2}P - \sigma \leqslant 0 \\ g_2 = \dfrac{A_2}{\sqrt{2}A_1^2 + 2A_1A_2}P - \sigma \leqslant 0 \\ g_3 = \dfrac{1}{A_1 + \sqrt{2}A_2}P - \sigma \leqslant 0 \end{cases}$$

where

$$A_1, A_2 \in [0, 1], l = 100cm, P = 2KN/cm^2, and\sigma = 2KN/cm^2$$

.

The results of the DSA on TBTP are shown in Table 10 and 11. The statistical results of the comparison algorithms are given in Table 10, and 11 presents the best solutions obtained by the DSA and other optimization algorithms. As shown, the optimal value of the DSA on TBTP is 263.8958434, which means when $x_1$, $x_2$, $g_1$, $g_2$, and $g_3$ are set to 0.788675136, 0.408248285, − 0.232790818, − 1.231270871, and − 1.001519946 respectively for the three-bar design problem. The convergence curves of this problem are shown in Fig. 13.

## 6.2 Sawmill operation problem

Assuming a company owns two sawmills and two forests. Duration of one project, each forest can produce up to 200 logs per day; the cost of transporting logs is estimated at \$10/km/log; and at least 300 logs are required per day. Table 12 shows the capacity of each of the mills (logs/day) and the distances between the forests and the mills (km). The goal is to **minimize the total daily cost of transporting logs and meet** the constraints on demand and factory capacity of the mills. The design problem is to determine how many logs to ship from Forest *one or two* to Mill *A or B* ($x_1$, $x_2$, $x_3$, $x_4$), as shown in Fig. 14.

The cost of transportation can be defined as follow:

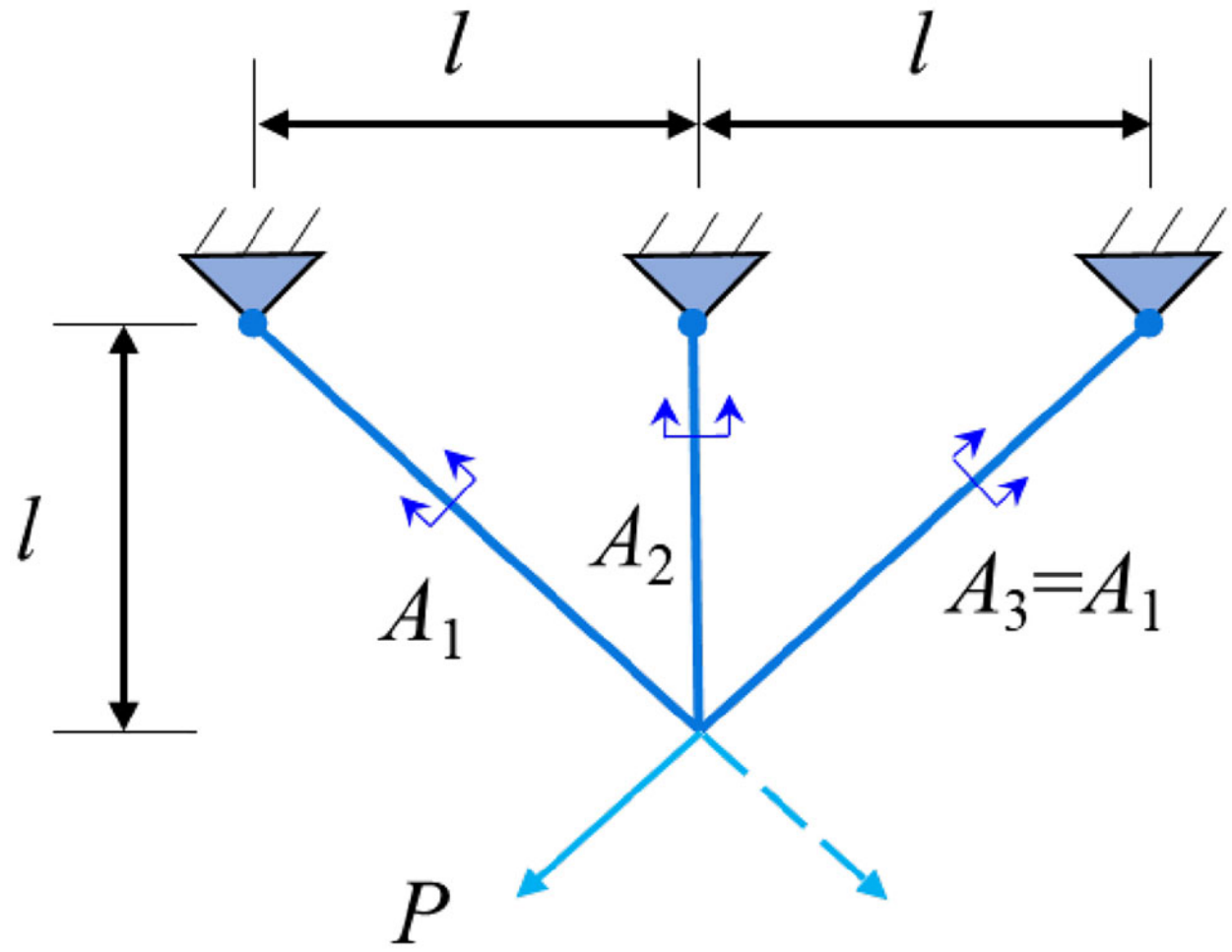

**Fig. 12** Three-bar truss structure

**Table 10** Comparison of statistical results of the DSA and other algorithms for solving the Three-bar truss problem

| Item | PSO | FA | CSO | GWO | SCA | MPA | AOA | DSA |
|---|---|---|---|---|---|---|---|---|
| Best | 270.6948 | 263.8968 | 263.8959 | 263.897 | 263.9446 | 263.8959 | 263.8961 | **263.8958** |
| Worst | 300.1633 | 263.9509 | 264.6476 | 263.939 | 282.8427 | 263.9515 | 263.9035 | 263.8959 |
| Mean | 282.9146 | 263.9069 | 264.0227 | 263.9074 | 269.7538 | 263.9074 | 263.8987 | 263.8959 |
| Std | 1.05E+01 | 1.66E−02 | 2.59E−01 | 1.52E−02 | 9.03E+00 | 2.01E-−02 | 2.44E−03 | 1.13E−05 |

Bold value indicates best value of the engineering constrained optimizationproblems from thecomparison algorithms

$$\min : f(x_1, x_2, x_3, x_4) = 10 \cdot (24x_1 + 20.5x_2 + 17.2x_3 + 10x_4) \tag{18}$$

$$\text{s.t}\begin{cases} g_1 = x_1 + x_2 - 240 \leqslant 0 \\ g_2 = x_3 + x_4 - 300 \leqslant 0 \\ g_3 = x_1 + x_3 - 200 \leqslant 0 \\ g_4 = x_2 + x_4 - 200 \leqslant 0 \\ g_5 = 300 - (x_1 + x_2 + x_3 + x_4) \leqslant 0 \end{cases}$$

where $x_1, x_2, x_3, x_4 \in [0, 200]$.

The results of the DSA on SOP are shown in Table 13 and 14. The statistical results of the comparison algorithms are listed in Table 13 and 14 presents the best solutions obtained by comparison methods. According to Table 14, the optimal value of the DSA on SOP is 37200.0053, which means when $x_1$, $x_2$, $x_3$, $x_4$, $g_1$, $g_2$, $g_3$, $g_4$, and $g_5$ are set to 1.20E−11, 1.63E−06, 100.0001, 199.9999, − 214.9374, − 281.9187, − 171.8508, − 196.9062, and 256.8561 respectively, the total cost of the SOP is the minimum. It can be concluded from Table 13 that the results obtained by the DSA are better than PSO, FA, CSO, GWO, SCA, and AOA, except MPA. The convergence capability of the proposed DSA and other algorithms is illustrated in Fig. 15.

### 6.3 Other engineering constrained problems

To further analyze the generalization performance of the DSA, four additional engineering constrained optimization problems, Tension spring design (TSD) [59, 66], Welded beam design (WBD) [18], Pressure vessel design (PVD) [59, 66], Speed reducer design (SRD) [60], were added to compare results. The parameters, dimension and other information of the engineering problem are shown in the following Table 15. The best optimization results of the DSA are compared with the existing advanced intelligent algorithms. The results of the comparison methods can be seen in Tables 16, 17, 18, 19 for details.

The results of the proposed DSA on TSD are listed in Table 16. Table 16 presents the best solutions obtained by the DSA and other optimization algorithms. The optimal value of the DSA on TSD is 0.012668, which means when $x_1$, $x_2$ and $x_3$ are set to 0.052068, 0.365900, and 10.770262 respectively for tension spring design. It can be concluded from Table 17 that the results obtained by the DSA are better than the comparison algorithms, except HGSO [33] and MPA [27].

The results of the proposed DSA on WBD are listed in Table 17. Table 17 shows the best solutions obtained by the comparison optimization algorithms. The optimal value of the DSA on WBD is 1.725555, which means when $x_1$, $x_2$, $x_3$ and $x_4$ are set to 0.205731, 3.475599, 9.036601, and 0.205731 respectively for welded beam design. It can be concluded from Table 17 that the results obtained by the DSA are better than the comparison algorithms, except AVOA [64].

The results of the proposed DSA on PVD are shown in Table 18. Table 18 presents the best solutions obtained by the DSA and other optimization algorithms. The optimal value of the DSA on PVD is 5885.374386, which means when $x_1$, $x_2$, $x_3$ and $x_4$ are set to 0.778189, 0.384659, 40.320642, and 199.985755 respectively for pressure vessel design. It can be concluded from Table 18 that the results obtained by the DSA are better than all the comparison algorithms.

The results of the proposed DSA on SRD are listed in Table 19. Table 19 presents the best solutions obtained by the DSA and other optimization algorithms. As shown, the optimal value of the DSA on SRD is 2996.403492, which means when $x_1$, $x_2$, $x_3$, $x_4$, $x_5$, $x_6$ and $x_7$ are set to 3.500006, 0.700000, 17.000000, 7.300490, 7.800000, 3.350216, and 5.286759 respectively for speed reducer design. The comparison results show that DSA is better than the optimization algorithm listed in Table 19.

### 6.4 Nodes optimization deployment of the WSN

Node optimization deployment [67–69] is the key point of the WSN for the development of the intelligent internet of things (IoT) and smart city, which can be applied to the smart agriculture, fisheries, animal husbandry, military, etc. Besides, node deployment also plays an important role in the field of environmental monitoring, which can used to collect the real time and accurate monitoring data. The node optimization deployment (NOD) problem [70] can be

**Table 11** The value of the decision variables and constraints in the best solution of the comparison algorithms

| Item | PSO | FA | CSO | GWO | SCA | MPA | AOA | DSA |
|---|---|---|---|---|---|---|---|---|
| $x_1$ | 0.728646364 | 0.787731811 | 0.788816646 | 0.788916075 | 0.79164305 | 0.78878101 | 0.788122965 | 0.788675136 |
| $x_2$ | 0.64602531 | 0.410925683 | 0.407848187 | 0.40757807 | 0.40034147 | 0.40794891 | 0.409812307 | 0.408248285 |
| $g_1$ | − 0.018676683 | − 2.05E−06 | − 3.15E−09 | −8.22E−06 | −0.0003251 | − 6.35E−09 | − 2.0000 | − 0.232790818 |
| $g_2$ | − 1.236507589 | − 1.461062784 | − 1.464556545 | −1.464867875 | −1.4732837 | − 1.464442 | − 2.0000 | − 1.231270871 |
| $g_3$ | − 0.782169094 | − 0.538939261 | − 0.535443458 | −0.535140345 | −0.5270414 | − 0.535558 | − 2.0000 | − 1.001519946 |
| $f$ | 270.6948451 | 263.8967705 | 263.8958586 | 263.8969697 | 263.944614 | 263.895852 | 263.896068 | **263.8958434** |

Bold value indicates best value of the engineering constrained optimizationproblems from thecomparison algorithms

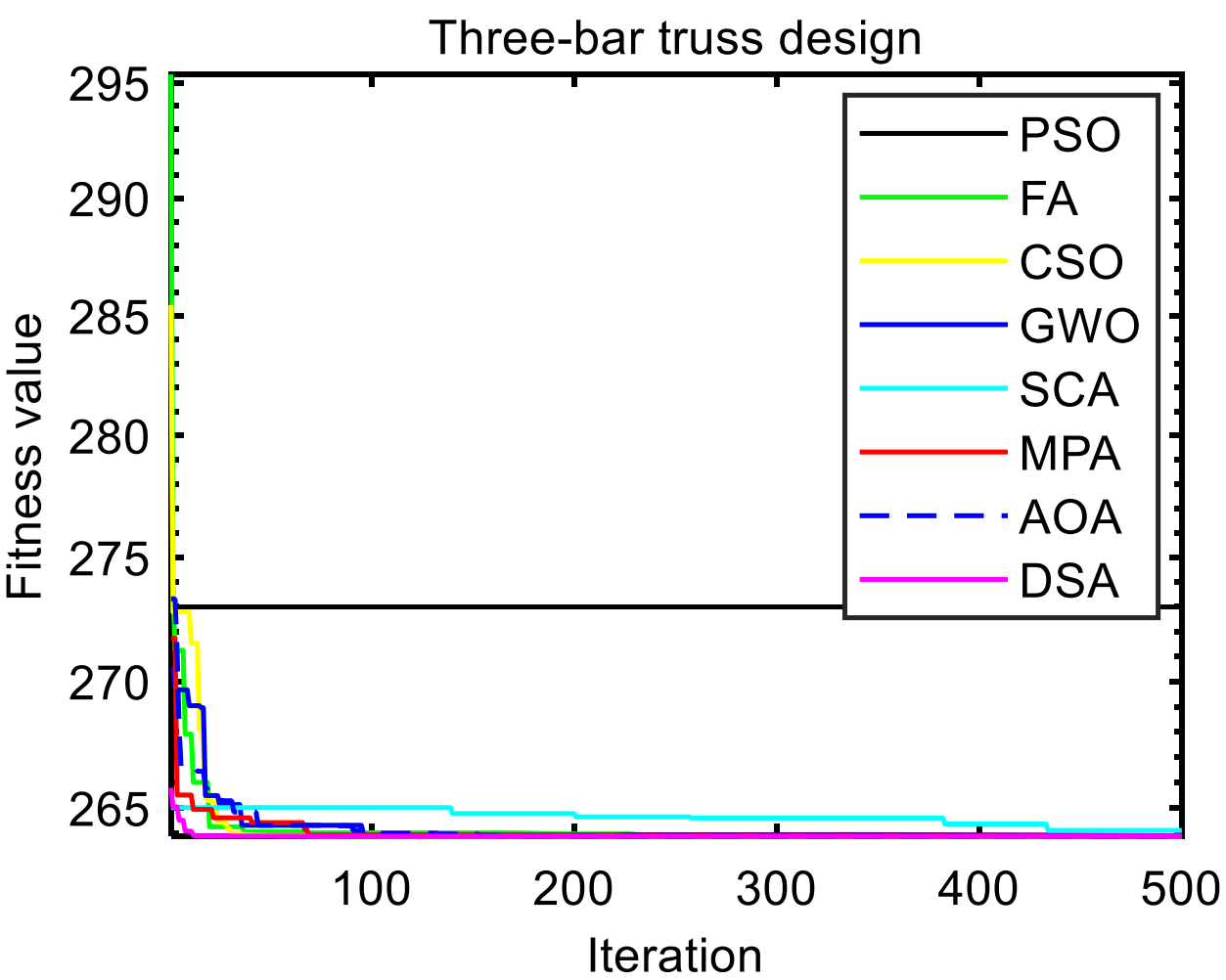

**Fig. 13** Convergence values curves of the Three-bar truss problem

**Table 12** Data for Sawmills

| Mill | Distance from forest one /km | Distance from forest two /km | Mill capacity per day/logs |
|---|---|---|---|
| A | 24.0 | 20.5 | 240 |
| B | 17.2 | 18.0 | 300 |

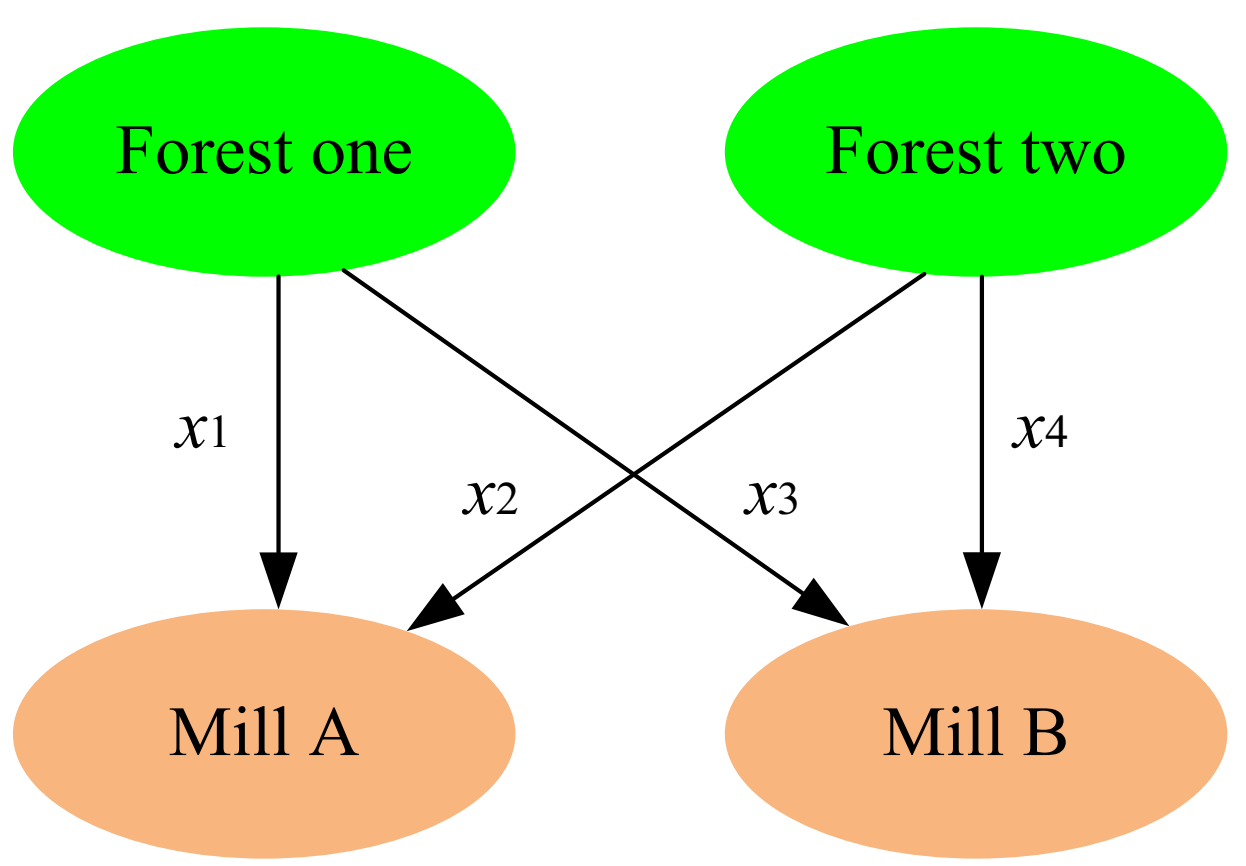

**Fig. 14** The Sawmill operation problem

regard as a constraint task. Supposing the length and width of the deployment area projected onto a two-dimensional plane are $L$ and $W$, and their unit is the meter (m). which is defined as:

**Table 13** Comparison of statistical results of the DSA and other algorithms for solving the Sawmill operation problem

| Item | PSO | FA | CSO | GWO | SCA | MPA | AOA | DSA |
| --- | --- | --- | --- | --- | --- | --- | --- | --- |
| Best | 43503.71 | 37212.54 | 37680.88 | 37407.78 | 37422.36 | **37200.01** | 37682.60 | **37200.01** |
| Worst | 55546.08 | 37638.43 | 39866.24 | 37790.57 | 58428.01 | 37200.13 | 43376.82 | 37723.63 |
| Mean | 50947.45 | 37288.92 | 38350.17 | 37588.82 | 40053.32 | 37200.06 | 39023.13 | 37384.72 |
| Std | 4.52E+03 | 1.27E + 02 | 6.38E + 02 | 1.15E + 02 | 6.47E + 03 | 3.40E-02 | 1.76E + 03 | 2.05E + 02 |

Bold value indicates best value of the engineering constrained optimizationproblems from thecomparison algorithms

$$\max\ Cov = \frac{\sum_{i=1,j=1}^{S,u_1 \times u_2} p(V_i, U_j)}{u_1 \times u_2} \times 100\%$$

$$s.t.\begin{cases} g_1 = \sum_{i=1,j=1}^{S,u_1 \times u_2} p(V_i, U_j) \geqslant 0 \\ g_2 = \sum_{i=1,j=1}^{S,u_1 \times u_2} p(V_i, U_j) - u_1 \times u_2 \geqslant 0 \\ g_3 = d(V_i, U_j) - R_s \geqslant 0 \\ g_4 = S - M \geqslant 0 \end{cases} \tag{19}$$

where $Cov$ denotes the node coverage percentage.$p(V_i, U_j)$ is the probability of the $j$-th monitoring point $U_j$ covering by the $i$-th sensor node $V_i$. $u_1, u_2$ indicate the monitoring point from x-axis and y-axis, respectively. $d(V_i, U_j)$ denotes the Euclidean distance from the $j$-th monitoring point $U_j$ to the $i$-th sensor node $V_i$. $R_s$ represents the node sensing radius, which satisfies $2R_s \leqslant R_c$, $R_c$ denotes the node communication radius. $S$ is the sensor node number in the monitoring area. $M$ is the deployment nodes number in theory. Notably, the binary perception model is used to calculate the coverage probability $p(V_i, U_j)$between the $j$-th monitoring point $U_j$and the $i$-th sensor node $V_i$, which is defined as

$$p(V_i, U_j) = \begin{cases} 0, & \text{if} \quad d(V_i, U_j) > R_s \\ 1, & \text{if} \quad d(V_i, U_j) \leq R_s \end{cases} \tag{20}$$

In addition to obtaining visual results intuitively through optimizing the coverage of nodes, the communication between nodes also needs to be considered. The Delaunay triangulation [71] is applied to analyze node connectivity, because the communication between nodes is generally based on the shortest distance strategy except for the sink nodes. Figure 16 shows the optimized node coverage results of the proposed DSA with different sensing radius. Besides, PSO, GWO, SCA, MPA, HBA are also utilized to verify the performance of the designed DSA for the NOD problem. The setting of parameters of the comparison algorithms are shown in Table 1. The $\beta$ and $C$ of the HBA are set 6 and 2, respectively. Notably, simulation parameters of the node deployment area are listed in Table 20. Besides, the comparison results of different algorithms are presented in the Table 21, which includes the node coverage percentage (%), optimization time (s), and the mean values.

In the deployment area of 100 m × 100 m, when the number of nodes is 20, different node sensing radius have a significant impact on the optimized coverage. This indicates that the larger the node sensing radius, the larger the range of sensing radius and the more data collected. The DSA optimized node coverage percentages are 88.68%,

**Table 14** The value of the decision variables and constraints in the best solution of the comparison algorithms

| Item | PSO | FA | CSO | GWO | SCA | MPA | AOA | DSA |
|---|---|---|---|---|---|---|---|---|
| $x_1$ | 14.0010 | 0.0726 | 0.7833 | 0.9979 | 1.1885 | 2.94E−08 | − 13.0210 | 1.20E−11 |
| $x_2$ | 45.1584 | 0.0000 | 3.8995 | 1.1851 | 0.0000 | 9.71E−05 | 13.0285 | 1.63E−06 |
| $x_3$ | 94.1700 | 100.0168 | 99.4557 | 99.2014 | 100.6120 | 100.0000 | 113.0214 | 100.0001 |
| $x_4$ | 146.8876 | 199.9223 | 195.8712 | 198.6269 | 198.3185 | 199.9999 | 186.9710 | 199.9999 |
| $g_1$ | − 135.2995 | − 239.9596 | − 233.1716 | − 235.8996 | − 240.0000 | − 239.9998 | − 240.4450 | − 214.9374 |
| $g_2$ | − 225.3532 | − 299.9013 | − 296.4594 | − 297.6044 | − 298.3476 | − 299.9996 | − 254.8587 | − 281.9187 |
| $g_3$ | − 125.2978 | − 199.9013 | − 193.7217 | − 199.7868 | −200.0000 | − 199.9994 | − 199.9993 | − 171.8508 |
| $g_4$ | − 78.2790 | − 199.8609 | − 194.8321 | − 195.6864 | − 200.0000 | − 199.9997 | − 84.1212 | − 196.9062 |
| $g_5$ | 120.6527 | 299.8609 | 289.6310 | 293.5040 | 298.3476 | 299.9994 | 255.3037 | 256.8561 |
| $f$ | 43503.7124 | 37212.5391 | 37680.8776 | 37407.7820 | 37422.3603 | 37200.0130 | 37682.6035 | 37200.0053 |

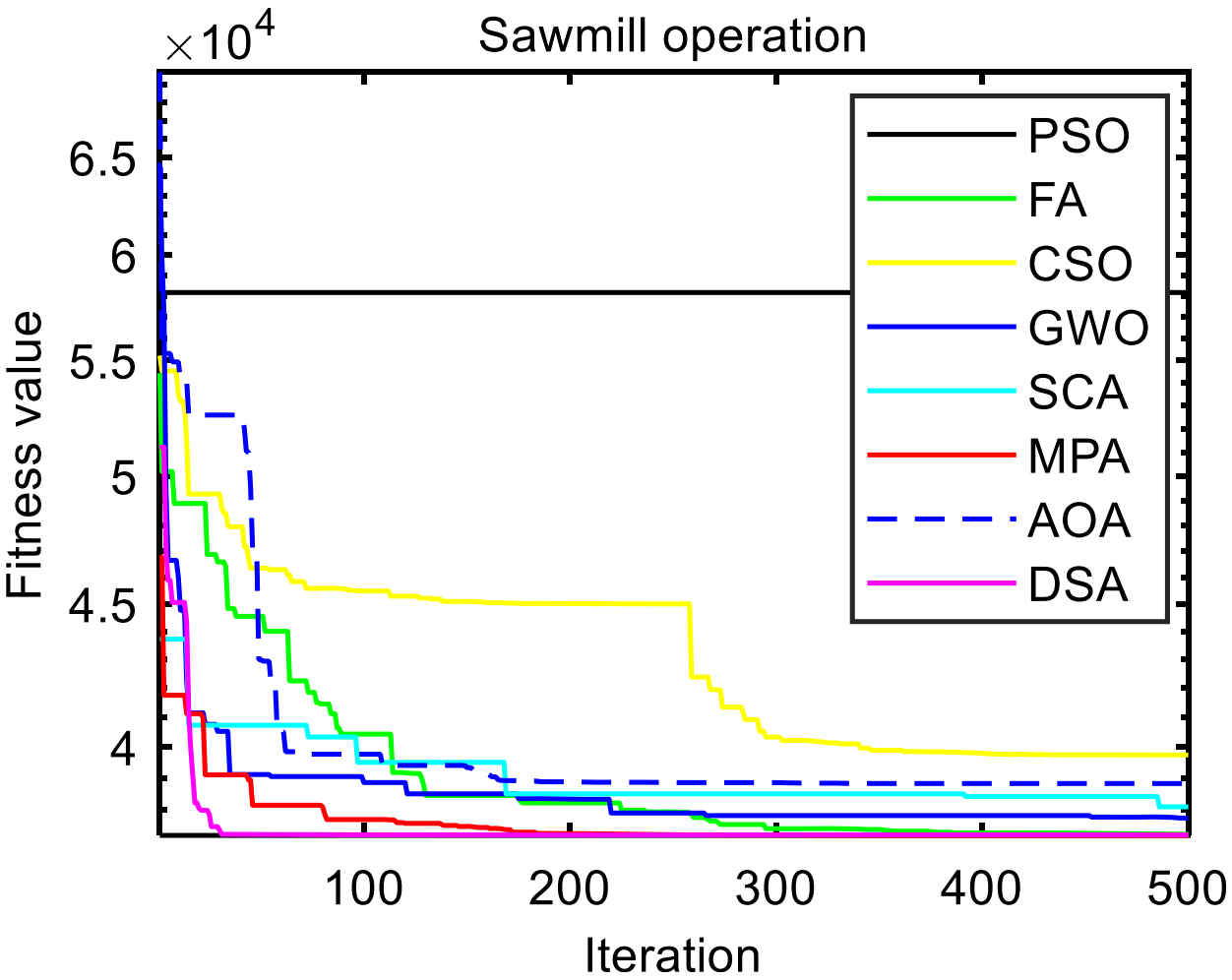

**Fig. 15** Convergence values curves of the Sawmill operation problem

92.82%, and 97.17%, respectively, and the time required to optimize node coordinates is 5.15 s, 5.34 s, and 4.98 s, respectively. From Fig. 16, when using random initial deployment (such as Fig. 16a, c, and e), not only will nodes exhibit centralized deployment, but the triangulation between nodes is also uneven, indicating poor regional coverage and connectivity of nodes. The deployment and Delaunay triangulation between nodes after DSA optimization are shown in Fig. 16 (b), (d), and (f), the optimized nodes have higher coverage and more uniform triangulation, indicating that communication between the optimized nodes will be smoother, thereby reducing power consumption. In summary, within the deployment area, when the number of nodes remains unchanged, using isomorphic sensors with a larger sensing radius can not only achieve better coverage, but also improve connectivity between network nodes, reduce communication power consumption between nodes, and increase network lifespan.

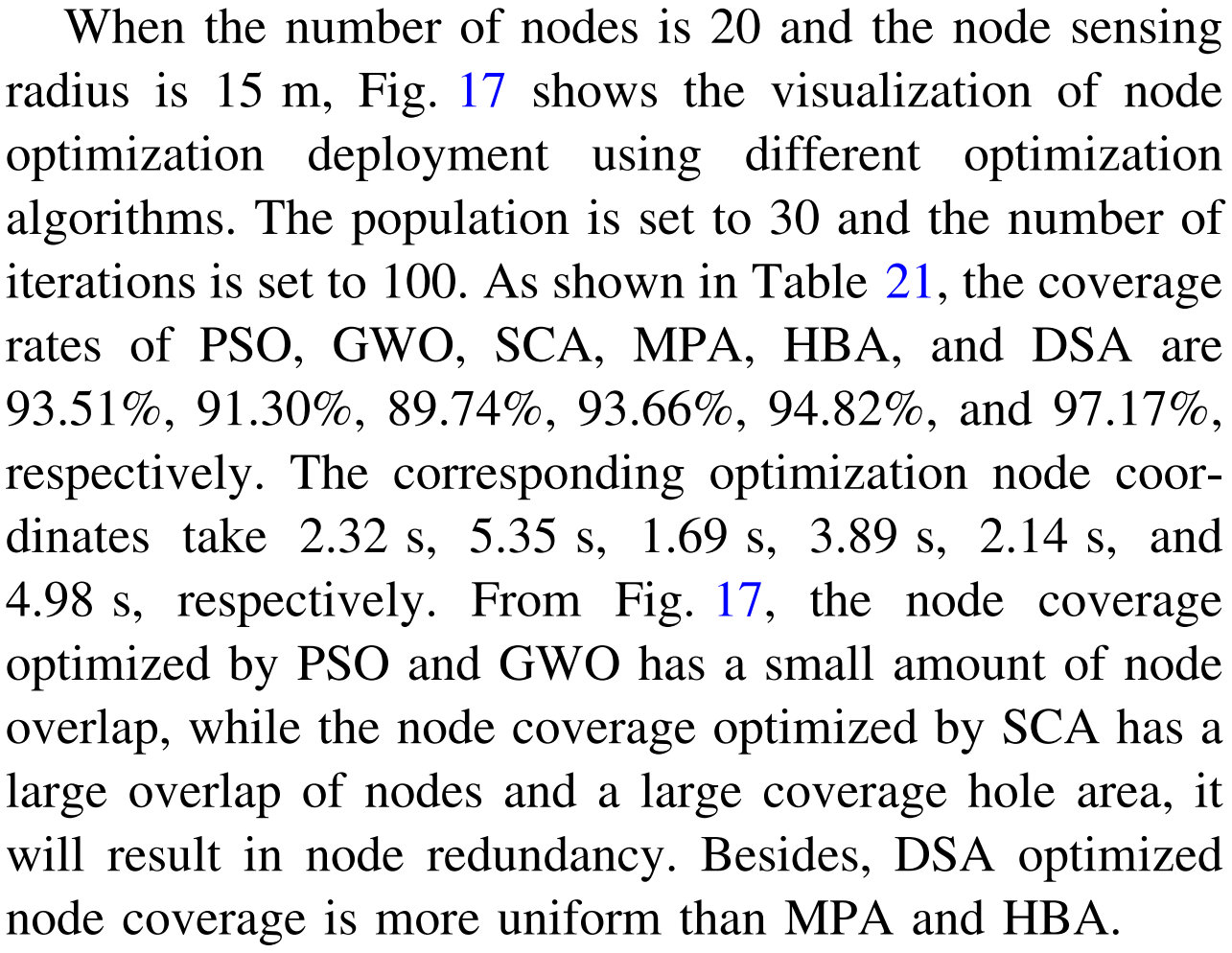

When the number of nodes is 20 and the node sensing radius is 15 m, Fig. 17 shows the visualization of node optimization deployment using different optimization algorithms. The population is set to 30 and the number of iterations is set to 100. As shown in Table 21, the coverage rates of PSO, GWO, SCA, MPA, HBA, and DSA are 93.51%, 91.30%, 89.74%, 93.66%, 94.82%, and 97.17%, respectively. The corresponding optimization node coordinates take 2.32 s, 5.35 s, 1.69 s, 3.89 s, 2.14 s, and 4.98 s, respectively. From Fig. 17, the node coverage optimized by PSO and GWO has a small amount of node overlap, while the node coverage optimized by SCA has a large overlap of nodes and a large coverage hole area, it will result in node redundancy. Besides, DSA optimized node coverage is more uniform than MPA and HBA.

From Table 21, for different node sensing radius, the coverage optimized by DSA is superior to other comparison algorithms, but node position optimization takes relatively long time. From the perspective of different node sensing radius, compared with PSO, the DSA node coverage corresponding to sensing radius of 13 m, 14 m, and 15 m has increased by 7.38%, 3.71%, and 3.66%, respectively. Compared with GWO, the coverage of DSA nodes with sensing radius of 13 m, 14 m, and 15 m increased by 6.82%, 6.46%, and 5.87, respectively. Compared with SCA, the coverage of DSA nodes with sensing radius of 13 m, 14 m, and 15 m increased by 8.31%, 5.01%, and 7.43%, respectively. Compared with MPA, the coverage of DSA nodes with sensing radius of 13 m, 14 m, and 15 m increased by 4.49%, 3.38%, and 3.51%, respectively. Compared with HBA, the coverage of DSA nodes with sensing radius of 13 m, 14 m, and 15 m increased by 3.27%, 2.21%, and 2.35%, respectively. From the mean coverage rate and optimization time of nodes' coordinates, the mean coverage rates of PSO, GWO, SCA, MPA, HBA, and DSA are 87.97%, 86.51%, 85.97%, 89.10%, 90.28%, and 92.89%, respectively. Although the optimization time of SCA is the least, the optimized node coverage is the lowest. The optimization time of DSA is only 0.12 s less than GWO, but the optimization performance of DSA is the best among the comparison algorithms. Therefore, DSA has better application prospects in WSN node coverage optimization problems, but there is still space for improvement in the performance of DSA.

**Table 15** The parameters, dimension, and other information of the four engineering constraint problems

| Problem name | Dim | parameters | Upper limit of the parameters | Lower limit of the parameters |
|---|---|---|---|---|
| Tension spring design | 3 | $x_1$, $x_2$, $x_3$ | [2.0 1.3 15.0] | [0.05 0.25 2.0] |
| Welded beam design | 4 | $x_1$, $x_2$, $x_3$, $x_4$ | [2 10 10 2] | [0.1 0.1 0.1 0.1] |
| Pressure vessel design | 4 | $x_1$, $x_2$, $x_3$, $x_4$ | [99 99 200 200] | [0 0 10 10] |
| Speed reducer design | 7 | $x_1$, $x_2$, $x_3$, $x_4$, $x_5$, $x_6$, $x_7$ | [3.6 0.8 28 8.3 8.3 3.9 5.5] | [2.6 0.7 17 7.3 7.8 2.9 5] |

**Table 16** The best value of Tension spring design of the comparison algorithms

| Algorithm | $x_1$ | $x_2$ | $x_3$ | optimal |
|---|---|---|---|---|
| PSO [33] | 0.0514 | 0.3577 | 11.6187 | 0.0127 |
| GA [59] | 0.051480 | 0.351661 | 11.632201 | 0.012704 |
| GSA [17] | 0.05028 | 0.32368 | 13.52541 | 0.01270 |
| GWO [18] | 0.0519 | 0.3627 | 10.9512 | 0.0127 |
| HGSO [33] | 0.0518 | 0.3569 | 11.2023 | 0.0126 |
| HHO [61] | 0.051796393 | 0.359305355 | 11.138859 | 0.012665443 |
| MPA [27] | 0.051724477 | 0.35757003 | 11.2391955 | 0.012665 |
| PO [40] | 0.05248 | 0.37594 | 10.24509 | 0.01267 |
| SSA [26] | 0.051207 | 0.345215 | 12.004032 | 0.0126763 |
| CPSO [62] | 0.051728 | 0.357644 | 11.244543 | 0.012674 |
| HBA [58] | 0.0506 | 0.3552 | 11.3730 | 0.01207 |
| εMAgES [60] | – | – | – | 1.67E+00 |
| iLSHADEε [60] | – | – | – | 1.67E+00 |
| DSA | 0.052068 | 0.365900 | 10.770262 | 0.012668 |

**Table 17** The best value of Welded beam design of the comparison algorithms

| Algorithm | $x_1$ | $x_2$ | $x_3$ | $x_4$ | Optimal |
|---|---|---|---|---|---|
| PSO [33] | 0.2157 | 3.4704 | 9.0356 | 0.2658 | 1.85778 |
| CS [25] | 0.182200 | 3.795100 | 9.998100 | 0.211100 | 1.946000 |
| GWO [18] | 0.2054 | 3.4778 | 9.0388 | 0.2067 | 1.7265 |
| GA [59] | 0.248900 | 6.173000 | 8.178900 | 0.253300 | 2.433116 |
| HGSO [33] | 0.2054 | 3.4476 | 9.0269 | 0.2060 | 1.7260 |
| GSA [17] | 0.182129 | 3.856979 | 10.0000 | 0.202376 | 1.879952 |
| AVOA [64] | 0.205730 | 3.470474 | 9.036621 | 0.205730 | 1.724852 |
| CPSO [62] | 0.202369 | 3.544214 | 9.048210 | 0.205723 | 1.728024 |
| HHO [61] | 0.204039 | 3.531061 | 9.027463 | 0.206147 | 1.73199057 |
| DSA | 0.205731 | 3.475599 | 9.036601 | 0.205731 | 1.725555 |

Figure 18 shows the node coverage optimization curves for algorithms such as PSO, GWO, SCA, MPA, HBA, and DSA, with 20 deployed nodes, 100 iterations, and a sensing radius of 13 m, 14 m, and 15 m, respectively. For different node sensing radius, DSA has achieved the optimal coverage percentage after 60 iterations, and it is significantly higher than other comparison algorithms around 40 iterations. MPA shows a clear upward trend in approximately 70 iterations. When the sensing radius $R_s$= 13 m, the curves of HBA, PSO, and SCA are relatively similar before 85 iterations, while SCA falls into local optima after 85 iterations. When the sensing radius $R_s$= 14 m, the node coverage of SCA optimization reached its optimal value around 40 times, which indicates that the SCA has fallen into local optima. When the sensing radius $R_s$= 15 m, the node coverage of HBA has significantly improved after 85 iterations and surpassed MPA. In summary, although DSA effectively improves network coverage, reduces coverage blind spots, and reduces coverage costs, the optimization time is more time-consuming than SCA, it indicates that the comprehensive performance of DSA in solving the node optimization coverage problem of WSN still needs to be improved.

**Table 18** The best value of Pressure vessel design of the comparison algorithms

| Algorithm | $x_1$ | $x_2$ | $x_3$ | $x_4$ | optimal |
|---|---|---|---|---|---|
| GA [59] | 0.812500 | 0.437500 | 42.097398 | 176.654050 | 6059.9463 |
| DE [63] | 0.812500 | 0.437500 | 42.098446 | 176.6360470 | 6059.701660 |
| GWO [18] | 0.8125 | 0.4345 | 42.089181 | 176.758731 | 6051.5639 |
| WOA [28] | 0.812500 | 0.437500 | 42.0982699 | 176.638998 | 6059.7410 |
| HHO [61] | 0.81758383 | 0.4072927 | 42.09174576 | 176.7196352 | 6000.46259 |
| MPA [27] | 0.8125 | 0.4375 | 42.098445 | 176.636607 | 6059.7144 |
| PO [40] | 0.7782 | 0.3847 | 40.3215 | 199.9733 | 5885.3997 |
| AOA [65] | 0.8303737 | 0.4162057 | 42.75127 | 169.3454 | 6048.7844 |
| AVOA [64] | 0.778954 | 0.3850374 | 40.360312 | 199.434299 | 5886.676593 |
| ϵMAgES [60] | – | – | – | – | 6.06E+03 |
| iLSHADEϵ [60] | – | – | – | – | 6.06E+03 |
| DSA | 0.778189 | 0.384659 | 40.320642 | 199.985755 | 5885.374386 |

**Table 19** The best value of Speed reducer design of the comparison algorithms

| Algorithm | $x_1$ | $x_2$ | $x_3$ | $x_4$ | $x_5$ | $x_6$ | $x_7$ | optimal |
|---|---|---|---|---|---|---|---|---|
| PSO [33] | 3.500 | 0.70 | 17 | 7.74 | 7.85 | 3.36 | 5.389 | 2998.12 |
| HS [13] | 3.520124 | 0.7 | 17 | 8.3 | 7.802354 | 3.366970 | 5.288719 | 3029.0020 |
| GSA [17] | 3.153 | 0.70 | 17 | 7.30 | 8.30 | 3.20 | 5.000 | 3040.10 |
| GWO [18] | 3.500 | 0.70 | 17 | 7.30 | 7.80 | 2.90 | 2.900 | 2998.83 |
| HGSO [33] | 3.498 | 0.71 | 17.02 | 7.67 | 7.810 | 3.36 | 5.289 | 2997.10 |
| SCA [32] | 3.508755 | 0.7 | 17 | 7.3 | 7.8 | 3.461020 | 5.289213 | 3030.563 |
| MFO [29] | 3.507524 | 0.7 | 17 | 7.302397 | 7.802364 | 3.323541 | 5.287524 | 3009.571 |
| εMAgES [60] | – | – | – | – | – | – | – | 2.99E+03 |
| iLSHADEε [60] | – | – | – | – | – | – | – | 2.99E+03 |
| DSA | 3.500006 | 0.700000 | 17.000000 | 7.300490 | 7.800000 | 3.350216 | 5.286759 | 2996.403492 |

## 7 Conclusion and future works

A novel SI optimization algorithm, named DSA is proposed inspired by the duck swarm in this study. Two rules are modeled from the finding food and foraging of the duck, which corresponds to the exploration and exploitation phases of the designed DSA, respectively. Where duck cooperation and competition are considered in details. Eighteen test functions from the CEC benchmark functions are utilized to evaluate the performance of the proposed algorithm in terms of exploration, exploitation, local optima avoidance, convergence, and diversity. The results show that DSA can provide highly competitive results compared to well-known optimization methods, such as PSO, FA, CSO, GWO, and SCA, and other recent algorithms like MPA and AOA. The superior exploitation, exploration, local optima avoidance ability of the DSA is confirmed by the results on different types of test functions. Moreover, the convergence analysis and population diversity analysis of the DSA confirm the convergence of this algorithm. Additionally, sensitivity analysis is used to access the performance of the proposed DSA. Furthermore, the DSA has high performance on the real-world engineering constrained optimization problems and the node optimization deployment task of the WSN.

For future work, we will further study on binary and multi-objective [72–74] versions of the DSA for solving the node computing and energy consumption optimization tasks of the IoT. The designed DSA is also can be used to

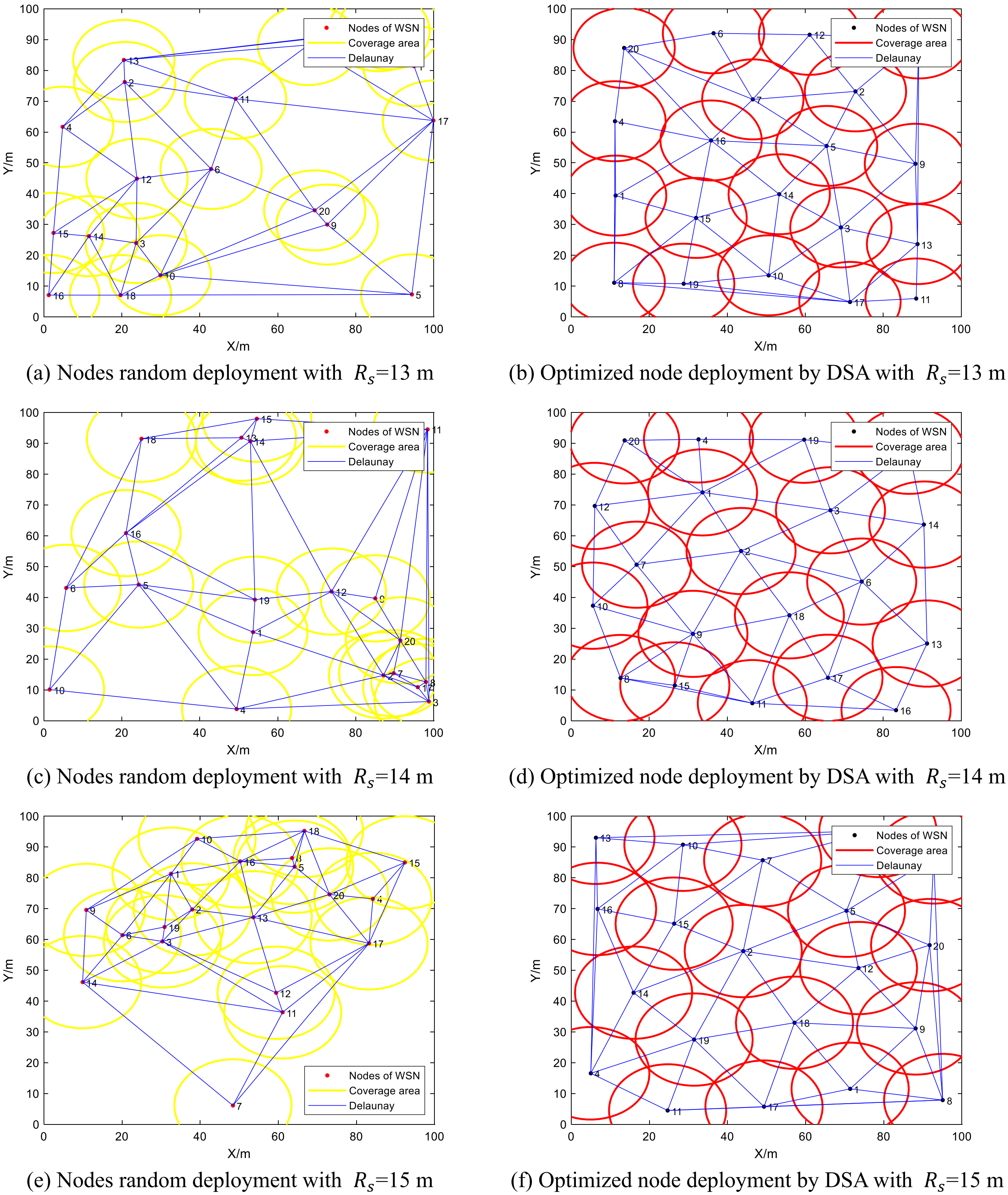

(a) Nodes random deployment with $R_s$=13 m

(b) Optimized node deployment by DSA with $R_s$=13 m

(c) Nodes random deployment with $R_s$=14 m

(d) Optimized node deployment by DSA with $R_s$=14 m

(e) Nodes random deployment with $R_s$=15 m

(f) Optimized node deployment by DSA with $R_s$=15 m

**Fig. 16** Visualization of node optimization deployment results using DSA

**Table 20** Simulation parameters of the node deployment area

| Parameters | Value |
|---|---|
| Deployment area | 100 m × 100 m |
| Node sensing radius $R_s$ | 13 m, 14 m, 15 m |
| Node communication radius $R_c$ | $2R_s$ |
| Number of nodes | 20 |
| Max iterations | 100 |
| Population number | 30 |

optimize the hyper-parameters of the deep learning models for the image classification [76, 77]. The feature selection problem and node location of the WSN also can be applied by the proposed inspired algorithm, which is considered the quantum theory [78].

**Table 21** The comparison results with different node sensing radius

| Method | Refs. | Cov/% $R_s$=13 m | Time/s | Cov/% $R_s$=14 m | Time/s | Cov/% $R_s$=15 m | Time/s | Cov/% (Mean) | Time/s (Mean) |
|---|---|---|---|---|---|---|---|---|---|
| PSO | [48] | 81.30 | 2.32 | 89.11 | 2.29 | 93.51 | 2.32 | 87.97 | 2.31 |
| GWO | [18] | 81.86 | 5.32 | 86.36 | 5.16 | 91.30 | 5.35 | 86.51 | 5.28 |
| SCA | [32] | 80.37 | **1.82** | 87.81 | **1.70** | 89.74 | **1.69** | 85.97 | **1.74** |
| MPA | [27] | 84.19 | 4.17 | 89.44 | 4.10 | 93.66 | 3.89 | 89.10 | 4.05 |
| HBA | [58] | 85.41 | 2.05 | 90.61 | 1.93 | 94.82 | 2.14 | 90.28 | 2.04 |
| DSA | – | **88.68** | 5.15 | **92.82** | 5.34 | **97.17** | 4.98 | **92.89** | 5.16 |

Bold values indicate optimal value of the comparison algorithms in this study

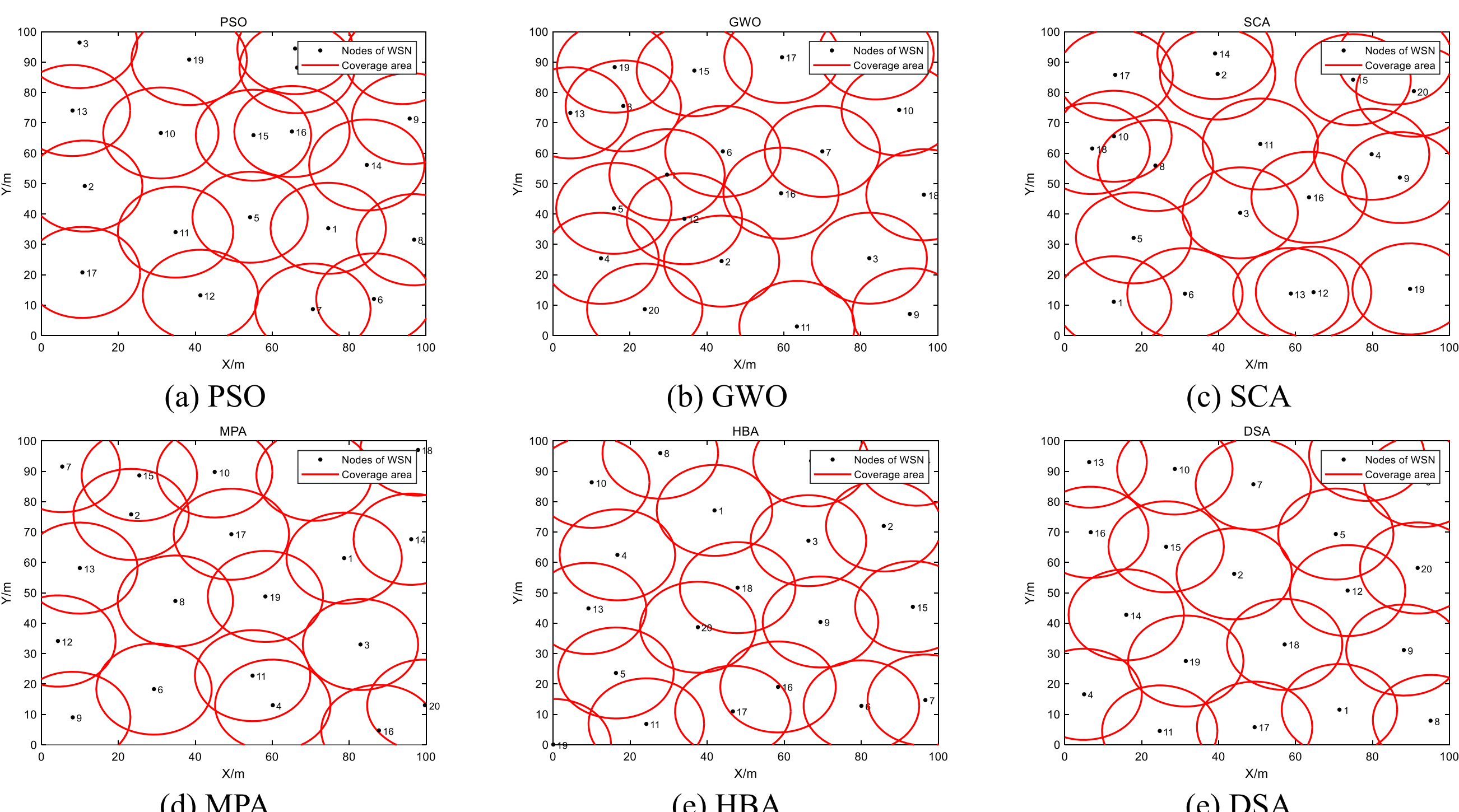

(a) PSO (b) GWO (c) SCA

(d) MPA (e) HBA (e) DSA

**Fig. 17** Visualization of node optimization deployment using different algorithms

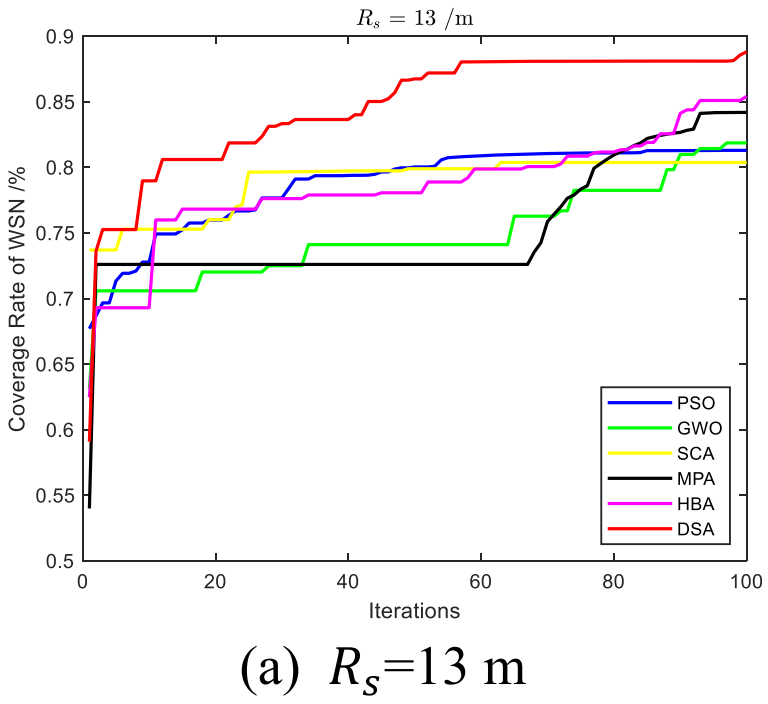

(a) $R_s$=13 m

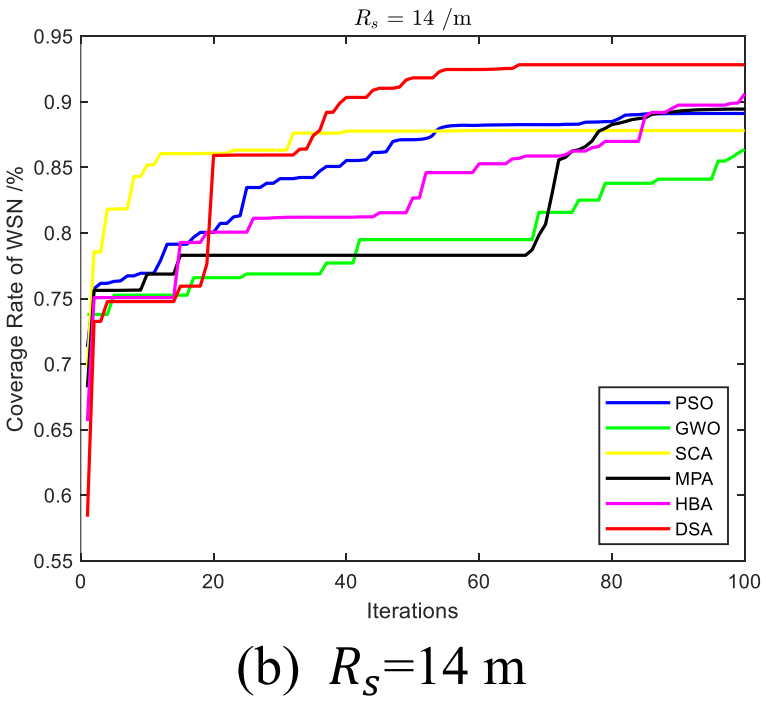

(b) $R_s$=14 m

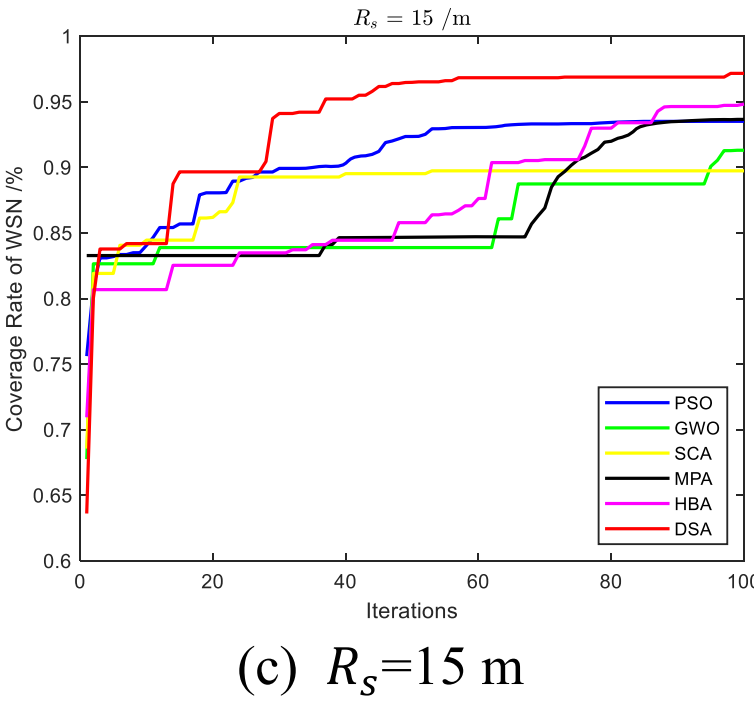

(c) $R_s$=15 m

**Fig. 18** Coverage curves of the comparison algorithms with different node sensing radius

**Author contributions** MZ: Conceptualization, Methodology, Writing—original draft, Writing—review and editing. GW: Supervision, Writing—review and editing.

**Funding** The authors have not disclosed any funding.

**Data availability** Enquiries about data availability should be directed to the authors.

## Declarations

**Conflict of interest** The authors declare that they have no conflict of interest.

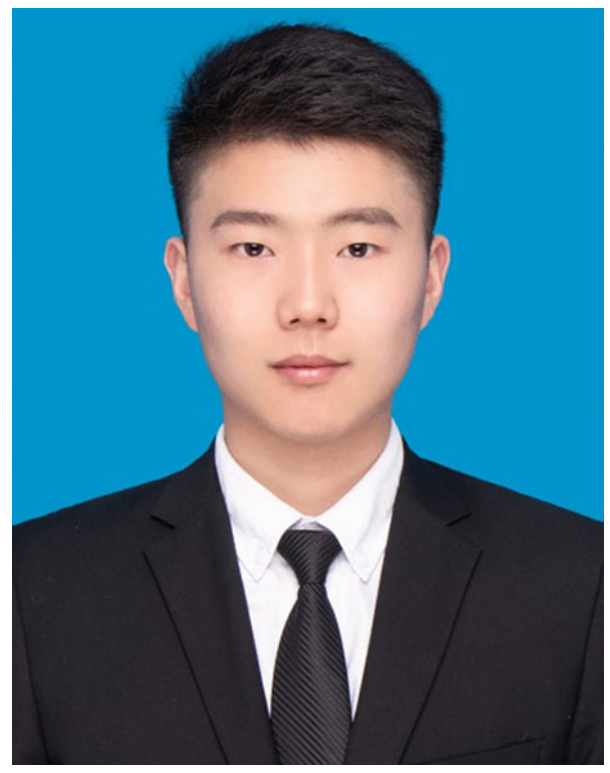

**Mengjian Zhang** received the M.S. degree in Control Science and Engineering from Guizhou University, Guiyang, China. He is pursuing the Eng.D. degree in Computer Technology from South China University of Technology, Guangzhou, China. His research interests include deep learning, medical image processing, and swarm intelligence algorithms and their applications.

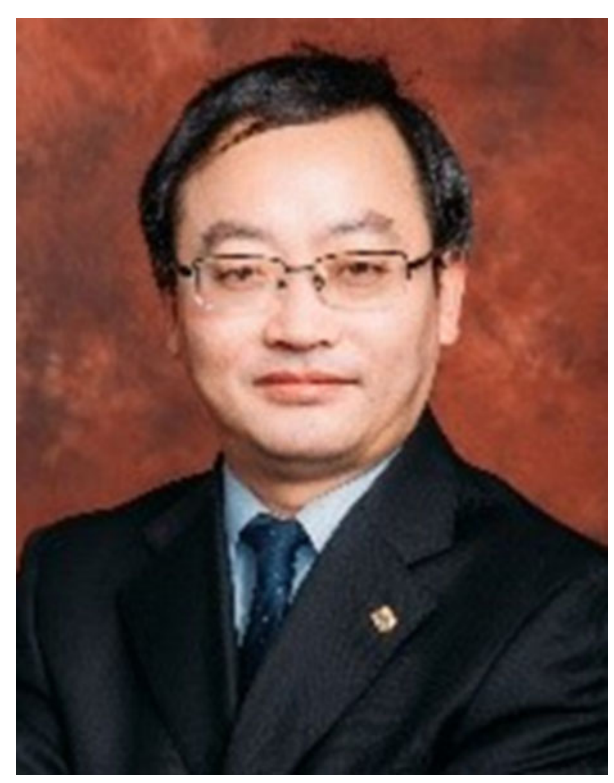

**Guihua Wen** received the Ph.D. degree in computer science and engineering from the South China University of Technology, Guangzhou, China. He is currently a Professor and DoctoralSupervisor at the School of Computer Science and Technology, South China University of Technology. His research interests include cognitive affective computing, machine learning, data mining, and intelligent medical methods.